\documentclass[runningheads]{llncs}

\usepackage{eccv}

\usepackage{eccvabbrv}  
\usepackage{graphicx}
\usepackage{booktabs}

\usepackage{multirow}
\usepackage{amsmath,amssymb}
\usepackage{dsfont}
\usepackage{xcolor}

\usepackage{hyperref}

\usepackage{orcidlink}

\newcommand{\vjepa}{V-JEPA2}
\newcommand{\vmae}{VideoMAEv2}
\newcommand{\dino}{DINOv2}
\newcommand{\rvm}{RVM}

\begin{document}

\title{Human-Level Accuracy, Non-Human Strategies:\\Revealing Model--Human Divergence in Video Physical Reasoning}

\titlerunning{Human-Level Accuracy, Non-Human Strategies}

\author{Fanhong Li\inst{1} \and
Shurui Zheng\inst{1} \and
Zi Yin\inst{1} \and
Junbo Cui\inst{2} \and
Lei Ji\inst{3} \and
Jia Liu\inst{1}\thanks{Corresponding author.}}

\authorrunning{F.~Li \etal}

\institute{
Tsinghua University, Beijing, China\\
\email{\{lifh21,zhengsr23,z-yin21\}@mails.tsinghua.edu.cn, liujiaTHU@tsinghua.edu.cn}
\and
ModelBest Inc., Beijing, China\\
\email{cuijb2000@gmail.com}
\and
Microsoft, Beijing, China\\
\email{leiji@microsoft.com}
}

\maketitle

\begin{abstract}
Video foundation models now reach human-level accuracy on physical-reasoning benchmarks, yet such tasks require predicting unobserved physical outcomes. Do these models perform human-like forward simulation, or do they exploit statistical regularities in visible scenes? Accuracy alone cannot distinguish these strategies.
We introduce a distributional evaluation framework that treats model seeds and human raters as populations, enabling comparison of consensus, uncertainty, and strategy. On the Physion benchmark, we evaluate three ViT-L architectures (\vjepa{}, \vmae{}, \dino{}). \vjepa{} narrows the accuracy gap to ${\sim}$1 percentage point ($73.2\%$ vs.\ $74.2\%$), yet model--human disagreement reaches $26.4\%$, far exceeding human--human disagreement ($4.8\%$), with substantially lower agreement ($\kappa \approx 0.48$ vs.\ $0.91$).
The divergence is scenario-structured: models outperform humans on geometric reasoning (linking, $+11.8$\,pp), while human advantages appear on causal-chain and other future-dynamics scenarios in the released split (e.g., dominoes, $-10.5$\,pp). Strategy fingerprinting confirms all three architectures share non-human strategies while none aligns with humans. Attribution analysis suggests that unobservable outcome features, rather than visible scene properties, predict this divergence, consistent with models relying more on scene-level statistical regularities than on explicit forward simulation, a systematic divergence that accuracy alone cannot reveal.
Code is available at \url{https://github.com/fanhong-li/model-human-divergence}.

\end{abstract}

\section{Introduction}
\label{sec:intro}

\begin{figure}[!t]
\centering
\includegraphics[width=\linewidth]{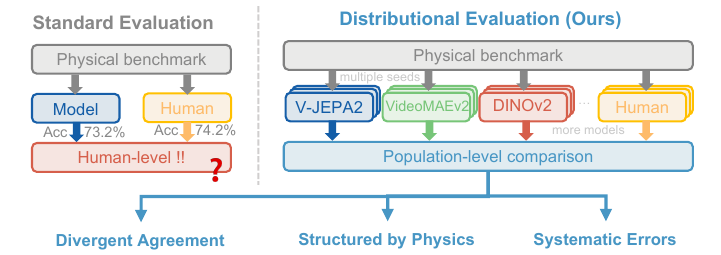}
\vspace{1mm}
\includegraphics[width=\linewidth]{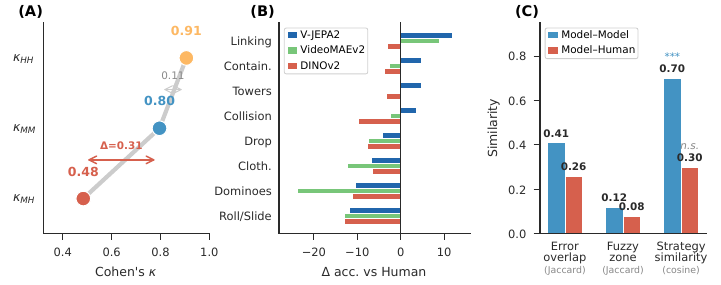}
\caption{\textbf{Overview.} \emph{Top:} Standard evaluation compares a single model's aggregate accuracy against human mean accuracy, masking systematic differences. Our distributional evaluation compares model populations (3 architectures $\times$ 8 seeds) against the full human rater population (${\sim}100$ raters $\times$ 1200 trials). \emph{Bottom:} Three key findings. (A)~Agreement hierarchy: model--model agreement ($\kappa_{MM}\!=\!0.80$) far exceeds model--human agreement ($\kappa_{MH}\!=\!0.48$), which in turn falls well below human--human agreement ($\kappa_{HH}\!=\!0.91$). (B)~Released-split scenario-specific accuracy gaps include a model advantage on linking ($+$11.8\,pp) and human advantages on rolling/sliding ($-$11.8\,pp) and dominoes ($-$10.5\,pp). (C)~Models share systematic errors with each other (error overlap Jaccard$\,{=}\,$0.41, strategy cosine$\,{=}\,$0.70) but not with humans (Jaccard$\,{=}\,$0.26, cosine$\,{=}\,$0.30). All Jaccard values are averages across model pairs.}
\label{fig:overview}
\end{figure}

Physical reasoning, \ie predicting how objects move, collide, and interact, is a core cognitive ability that develops early in infancy~\cite{Spelke1992,Battaglia2013,Kubricht2017}. A distinguishing feature of physical reasoning is that it requires predicting unobserved outcomes: given partial observation of a scene, humans mentally simulate forward in time to anticipate what will happen next~\cite{Battaglia2013}. Recent self-supervised video models have made striking progress on physical reasoning benchmarks. V-JEPA~\cite{Bardes2024} reaches 98\% zero-shot accuracy on violation-of-expectation tasks~\cite{Garrido2025}, and \vjepa{}~\cite{Assran2025} matches human-level performance on contact prediction~\cite{Bear2021}. But do these results reflect human-like forward simulation, or a qualitatively different strategy that extracts statistical regularities from visible scenes without simulating unobserved dynamics?

A single accuracy number cannot distinguish between these possibilities. Two systems can achieve identical scores while making errors on entirely different samples~\cite{Geirhos2020,GeirhosError2020}, implying fundamentally different underlying strategies. When a physical-reasoning benchmark contains a mix of scenarios, some solvable from geometric cues alone, others requiring forward simulation of gravitational dynamics or causal chains, aggregate accuracy conflates performance on both types, masking systematic differences in \emph{which} scenarios each system solves. Distinguishing these possibilities requires comparing not just how often models and humans are correct, but \emph{on which samples} they agree, how confidently they respond, and whether their errors cluster in the same places.

\paragraph{Our Approach.}
We address this with a distributional evaluation framework that treats model seeds and human raters as \emph{populations} rather than point estimates. For models, multiple random probe initializations on the same frozen encoder produce a population of predictions per sample; for humans, ${\sim}100$ independent raters per stimulus provide the reference population. This enables three levels of analysis: consensus comparison via inter-rater agreement metrics, uncertainty decomposition via entropy analysis, and strategy fingerprinting via physical-feature attribution. Critically, by evaluating three ViT-L architectures (\vjepa{}~\cite{Assran2025}, \vmae{}~\cite{Wang2023}, \dino{}~\cite{Oquab2024}) under a unified protocol, identical backbone size, frame sampling, and probe architecture, we can test whether divergence from human behavior is architecture-specific or shared across the evaluated frozen representations.

\paragraph{Key Findings.}
Applying this framework to the Physion benchmark~\cite{Bear2021} reveals converging evidence at three levels.
\emph{Existence:} \vjepa{} achieves $73.2\%{\pm}0.95\%$, within 1\,pp of human accuracy ($74.2\%$), yet model--human disagreement ($26.4\%$) is $5.5{\times}$ higher than human--human disagreement ($4.8\%$; $\kappa\!=\!0.48$ vs.\ $0.91$).
\emph{Structure:} divergence is scenario-structured: models outperform humans on geometric scenarios (linking $+11.8$\,pp), while human advantages appear on causal-chain and other future-dynamics scenarios in the released split (dominoes $-10.5$\,pp; rolling/sliding $-11.8$\,pp), consistently across all three architectures.
\emph{Source:} in our attribution analysis, observable scene properties carry little accessible signal for where models and humans diverge, whereas unobservable outcome features do; this pattern is consistent with divergence tied to forward simulation rather than scene encoding alone.

\paragraph{Contributions.}
\begin{itemize}
  \item A \emph{distributional evaluation framework} that treats model seeds and human raters as populations, enabling consensus, uncertainty, and strategy comparison beyond aggregate accuracy (\S\ref{sec:method}).
  \item A three-level characterization of model--human divergence on physical reasoning, from existence (systematic disagreement despite matched accuracy) through structure (released-split scenario differences) to source (unobservable outcome features predict divergence better than observable features in the attribution analysis), established across three ViT-L architectures under fair comparison (\S\ref{sec:results}, \S\ref{sec:diagnostic}).
  \item Evidence that, for the evaluated frozen ViT-L representations and supervised probes, divergence is more strongly associated with unobserved physical outcomes than with visible-scene features, a limitation that accuracy-based evaluation cannot detect.
\end{itemize}

\section{Related Work}
\label{sec:related}

\paragraph{Physical Reasoning Benchmarks.}
Physion~\cite{Bear2021} introduced eight physical scenarios with per-stimulus human judgments (${\sim}100$ raters), establishing frozen-encoder probing for binary contact prediction. Physion++~\cite{Tung2024} extends this to latent physical properties. Other benchmarks include IntPhys~\cite{Riochet2022}, CLEVRER~\cite{Yi2020}, and CoPhy~\cite{Baradel2020}. Garrido~\etal~\cite{Garrido2025} showed that V-JEPA acquires intuitive physics through self-supervised pretraining but revealed weaknesses on inter-object interactions. These benchmarks provide rich human response data, yet prior work uses only aggregate accuracy for model--human comparison, without examining whether model performance reflects the same reasoning mechanisms as humans.

\paragraph{Simulation vs.\ Pattern Matching in Physical Reasoning.}
Cognitive science distinguishes two accounts of physical reasoning. The \emph{simulation} account~\cite{Battaglia2013} proposes that humans predict physical outcomes by running approximate, probabilistic simulations of scene dynamics, an ``intuitive physics engine'' analogous to game-engine physics. This view is supported by systematic biases in human judgments that match simulation noise~\cite{Kubricht2017}, and has been formalized as a core component of human-like reasoning~\cite{Lake2017}. The alternative, \emph{pattern matching}, holds that many physical judgments can be made from static or early-dynamic cues without forward simulation. In deep learning, evidence is mixed: Piloto~\etal~\cite{Piloto2022} showed that a graph network trained on physical dynamics develops violation-of-expectation responses, while Garrido~\etal~\cite{Garrido2025} found that self-supervised video models acquire some intuitive physics but fail on inter-object interaction, a capacity closely tied to forward simulation. These are scattered observations; no prior work has systematically tested, in a controlled setting, whether the locus of model--human divergence lies in processing visible information or in reasoning about unobserved futures.

\paragraph{Beyond-Accuracy and Distributional Evaluation.}
Geirhos~\etal~\cite{Geirhos2021,GeirhosError2020} demonstrated that vision models with similar accuracy exhibit fundamentally different error patterns, advocating for error consistency analysis. This ``beyond accuracy'' perspective has been applied to image classification~\cite{Geirhos2021} and robustness evaluation~\cite{Hendrycks2021}, but not to physical reasoning. Complementarily, psychometric traditions, Item Response Theory~\cite{Embretson2000} and inter-rater reliability (Fleiss' $\kappa$, Krippendorff's $\alpha$), provide tools for comparing agent populations. Recent work applies these ideas to LLM evaluation~\cite{Burnell2023} and human-AI complementarity~\cite{Bansal2021}. We unify both lines: error consistency analysis to quantify \emph{what} diverges and population-level agreement metrics to quantify \emph{how much}, applied for the first time to physical reasoning where the simulation-vs-cue-based-reasoning question gives these tools a specific theoretical target.

\paragraph{Probing Frozen Representations.}
Linear probing~\cite{Alain2017,Hewitt2019,Belinkov2022} is standard for representation evaluation. Concurrent with our work, Joseph~\etal~\cite{Joseph2026} identify a ``Physics Emergence Zone'' at intermediate layers of \vjepa{}, a pattern we independently confirm. While their work focuses on \emph{what} physical information is represented, we address \emph{whether} model and human populations use it in the same way.

\section{Distributional Evaluation Framework}
\label{sec:method}

Standard evaluation compares a single model prediction against ground truth, yielding accuracy. Our approach treats both model instantiations and human raters as \emph{populations}, enabling richer comparisons at the consensus, uncertainty, and strategy levels.

\subsection{Population Representation}
\label{sec:notation}

Let $N\!=\!1200$ denote test samples, $K\!=\!8$ model seeds (random probe initializations on a frozen encoder), and $H_i \!\approx\! 100$ human raters for sample~$i$. We define:

\begin{itemize}
    \item $\mathbf{M} \in \{0,1\}^{N \times K}$: model predictions, where $M_{ik}\!=\!1$ if seed $k$ predicts contact for sample~$i$.
    \item $\mathbf{R} \in \{0,1\}^{N \times H_{\max}}$: human responses ($H_i\!\approx\!100$ varies by sample; unused entries are masked).
    \item $y \in \{0,1\}^N$: ground-truth labels.
\end{itemize}

\noindent From these, we derive population-level consensus:
\begin{equation}
p_m^{(i)} = \frac{1}{K}\sum_{k=1}^{K} M_{ik}, \qquad
p_h^{(i)} = \frac{1}{H_i}\sum_{j=1}^{H_i} R_{ij},
\label{eq:consensus}
\end{equation}
representing the fraction of each population predicting contact. The majority-vote predictions are $\hat{y}_m^{(i)} = \mathds{1}\!\left[p_m^{(i)} > 0.5\right]$ and $\hat{y}_h^{(i)} = \mathds{1}\!\left[p_h^{(i)} > 0.5\right]$.

\subsection{Three-Level Agreement Hierarchy}
\label{sec:three_level}

We decompose agreement into three levels, each computed via split-half bootstrap (1,000 iterations) to produce confidence intervals:

\paragraph{Level 1: Human--Human.} Randomly split $H$ raters into two halves; each half produces a majority vote. Cohen's $\kappa$ between the two halves, averaged over bootstrap iterations, yields $\kappa_{\mathrm{HH}}$.

\paragraph{Level 2: Model--Model.} Randomly split $K$ seeds into two halves; each half produces a majority vote. Cohen's $\kappa$ between the two halves yields $\kappa_{\mathrm{MM}}$.

\paragraph{Level 3: Model--Human.} Cohen's $\kappa$ between the model majority vote $\hat{y}_m$ and the human majority vote $\hat{y}_h$ yields $\kappa_{\mathrm{MH}}$. To obtain a confidence interval comparable to the other two levels, we bootstrap $\kappa_{\mathrm{MH}}$ by jointly resampling seeds and raters (1,000 iterations).

\noindent The expected ordering $\kappa_{\mathrm{HH}} \gg \kappa_{\mathrm{MM}} \gg \kappa_{\mathrm{MH}}$ holds if models form a coherent population that differs systematically from the human population. Deviations from this ordering are diagnostic: $\kappa_{\mathrm{MM}} \approx \kappa_{\mathrm{HH}}$ would suggest model seeds are as reliable as human raters; $\kappa_{\mathrm{MM}} \approx \kappa_{\mathrm{MH}}$ would suggest model internal disagreement fully accounts for model--human divergence.

\subsection{Uncertainty Decomposition}
\label{sec:entropy_method}

For each sample, we compute the binary entropy of each population's consensus:
\begin{equation}
\mathcal{H}_m^{(i)} = H(p_m^{(i)}), \qquad \mathcal{H}_h^{(i)} = H(p_h^{(i)}),
\label{eq:entropy}
\end{equation}
where $H(p) = -p\log_2 p - (1{-}p)\log_2(1{-}p)$. This induces a four-quadrant classification at threshold $\tau$ (we use $\tau\!=\!H(0.25)\!\approx\!0.81$ bits, corresponding to 25\%/75\% consensus):

\begin{itemize}
    \item \textbf{Q1} ($\mathcal{H}_m < \tau$, $\mathcal{H}_h < \tau$): Both populations certain, ``easy'' samples.
    \item \textbf{Q2} ($\mathcal{H}_m \geq \tau$, $\mathcal{H}_h < \tau$): Model uncertain, humans certain; model blind spots.
    \item \textbf{Q3} ($\mathcal{H}_m < \tau$, $\mathcal{H}_h \geq \tau$): Model certain, humans uncertain; confident divergence.
    \item \textbf{Q4} ($\mathcal{H}_m \geq \tau$, $\mathcal{H}_h \geq \tau$): Both uncertain, genuinely ambiguous.
\end{itemize}

\noindent The asymmetry ratio $A = |Q2|/|Q3|$ quantifies the directionality of information access: $A \gg 1$ indicates systematic model blind spots, while $A \ll 1$ indicates models are confidently committed where humans are uncertain.

\subsection{Strategy Fingerprinting}
\label{sec:fingerprint_method}

To characterize \emph{how} each population makes decisions, we extract physical features from the Physion HDF5 simulation data organized into three temporal layers: \textbf{Layer~A} (static scene properties: mass, friction, object counts, geometry), \textbf{Layer~B} (observable dynamics up to the cutoff frame: speeds, kinetic energies, collision counts, trajectory statistics), and \textbf{Layer~C} (unobservable outcome features: post-cutoff contact timing, causal chain length, prediction horizon). We use this layered framework at two levels of granularity depending on the analysis goal.

For \emph{strategy fingerprinting} (\S\ref{sec:structured}), we use a 37-feature core subset spanning all three layers (6A\,+\,24B\,+\,7C), selected to capture representative physical quantities at coarse granularity. For each population, the regression target is the length-$N$ correctness vector over the same 1,200 stimuli ($1$ if majority vote matches ground truth, $0$ otherwise), so model and human fingerprints are estimated from directly aligned item-level behavior. We train Ridge regression (L2-penalized least squares) to predict this per-sample correctness vector from the 37 physical features. We choose Ridge over logistic regression because the MSE objective yields coefficient vectors reflecting each feature's marginal linear contribution to accuracy variance, providing a more graded signal for pairwise cosine comparison than log-likelihood-optimized coefficients. The resulting $\boldsymbol{\beta} \in \mathbb{R}^{37}$ serves as a ``strategy fingerprint,'' and pairwise cosine similarity between fingerprints quantifies strategy alignment. For \emph{feature attribution} (\S\ref{sec:diagnostic}), we use the full 89-feature set (19A\,+\,47B\,+\,23C), a systematic expansion that adds zone properties, probe/force information, collision detail breakdown, and composite outcome indices. The expanded set enables layer-wise attribution of model--human divergence. The 37-feature subset has direct semantic equivalents for 32/37 features in the 89-feature set; the five without equivalents are camera-relative or derivative quantities excluded from the systematic extraction (full correspondence table in supplementary material).

We note that strategy fingerprints capture \emph{correlational} structure between physical features and accuracy patterns, not causal mechanisms; we use ``strategy'' as shorthand for this statistical signature.

\section{Experimental Setup}
\label{sec:setup}

\subsection{Dataset: Physion}
\label{sec:dataset}

Physion~\cite{Bear2021} consists of 8 physical scenarios (\emph{collision}, \emph{containment}, \emph{dominoes}, \emph{drop}, \emph{linking}, \emph{rolling/sliding}, \emph{towers}, \emph{clothiness}), each containing 150 test samples (1,200 total) and ${\sim}$1,000 training samples (8,005 total). Each 5-second video is rendered from a physics simulator (ThreeDWorld); the task is binary contact prediction: will two designated objects come into contact? Human judgments from ${\sim}100$ Amazon Mechanical Turk raters per stimulus provide the reference population, totaling 120,450 raw responses across all 1,200 test samples.

The HDF5 simulation files provide complete physical state trajectories (per-frame object positions, velocities, collisions, physical properties), enabling extraction of the layered physical features used for strategy fingerprinting and attribution analysis (\S\ref{sec:fingerprint_method}).

\subsection{Models and Fair Comparison Protocol}
\label{sec:models}

A central design goal is \emph{fair cross-model comparison}. Prior work compared models with different ViT sizes (Base/Large/Giant), frame rates, and probe architectures, confounding these factors with model quality. We standardize:

\paragraph{Unified backbone.} All models use \textbf{ViT-Large} (${\sim}307$M parameters):
\begin{itemize}
    \item \textbf{\vjepa{}}~\cite{Assran2025}: video joint-embedding predictive architecture.
    \item \textbf{\vmae{}}~\cite{Wang2023}: video masked autoencoder.
    \item \textbf{\dino{}}~\cite{Oquab2024}: image self-distillation, applied per-frame.
\end{itemize}

\paragraph{Unified evaluation.} 8 input frames, fractional frame step 5.75 (${\sim}$1.5s temporal coverage, matching the human observation window), same E2E attentive probe architecture (depth 4, 16 heads), identical hyperparameter search grid (5 learning rates $\times$ 2 weight decay values $=$ 10 probe configurations per seed), 20 training epochs. Best head selected by training accuracy (standard Physion protocol). A supplementary head-selection audit shows that choosing the test-oracle head would change accuracy by only 0.36pp, so probe selection cannot explain the much larger model--human disagreement.

\paragraph{Multi-seed evaluation.} Each model is trained with \textbf{8 random seeds} (probe initialization only; encoders remain frozen). This enables population-level analysis: each seed produces an independent prediction vector over all 1,200 test samples.

\subsection{Human Population Data}
\label{sec:human_data}

For each test sample, ${\sim}100$ independent raters provide binary contact predictions. We use the full response matrix $\mathbf{R} \in \{0,1\}^{1200 \times H}$ (where $H$ varies by scenario, averaging ${\sim}100$) for all distributional analyses. The human majority vote serves as the reference prediction; the response fraction $p_h^{(i)}$ quantifies human consensus per sample. All human data are drawn from the published Physion benchmark~\cite{Bear2021}; the original data collection protocol, including IRB approval and informed consent procedures, is described therein.

\section{Results: Convergent Scores, Divergent Solutions}
\label{sec:results}

\subsection{Surface Equivalence}
\label{sec:surface}

\begin{table}[t]
\centering
\caption{Cross-model comparison on Physion (all ViT-L, unified protocol, E2E attentive probes). \vjepa{} significantly outperforms alternatives ($p < 0.001$, McNemar). All models show $>$\,$5{\times}$ the disagreement rate of human--human baseline (4.8\%).}
\label{tab:crossmodel}
\small\setlength{\tabcolsep}{4pt}
\begin{tabular}{lccccc}
\toprule
Model & Seeds & Test Acc. & Train Acc. & Disagr. & $\kappa_{\mathrm{MH}}$ \\
\midrule
\vjepa{}      & 8 & $\mathbf{73.16{\pm}0.95}$ & $90.60{\pm}0.56$ & $26.4$ & $0.484$ \\
\vmae{}       & 8  & $67.65{\pm}0.72$ & $85.41{\pm}0.47$ & $30.0$ & $0.397$ \\
\dino{}       & 8  & $66.90{\pm}0.73$ & $91.96{\pm}0.38$ & $30.2$ & $0.399$ \\
Human         & ${\sim}$100 & $74.2$ & --- & $4.8$ & $0.906$ (HH) \\
\bottomrule
\end{tabular}
\end{table}

\noindent\textbf{Finding 1:} \emph{\vjepa{} achieves $73.2\%{\pm}0.95\%$, within 1.0pp of human accuracy ($74.2\%$), but this apparent convergence is misleading.}

Table~\ref{tab:crossmodel} summarizes overall accuracy. \vjepa{} leads by ${\sim}5$--$6$pp over alternatives (all differences significant). \dino{} shows the largest train--test gap (25.1pp), suggesting partial memorization, though its per-scenario error patterns remain qualitatively consistent with the other models (Table~\ref{tab:scenario}). The \vjepa{}--human gap (1.0pp) might suggest human-level reasoning; the remainder of this section shows why this conclusion is premature.

\paragraph{The agreement hierarchy.}
The three-level agreement hierarchy (\S\ref{sec:three_level}), estimated via split-half bootstrap, yields:

\begin{equation}
\underbrace{\kappa_{\mathrm{HH}} = 0.906\;{\scriptstyle[.89,.92]}}_{\text{human--human}} \;\gg\; \underbrace{\kappa_{\mathrm{MM}} = 0.796\;{\scriptstyle[.77,.82]}}_{\text{model--model}} \;\gg\; \underbrace{\kappa_{\mathrm{MH}} = 0.484\;{\scriptstyle[.44,.49]}}_{\text{model--human}}
\label{eq:hierarchy}
\end{equation}

\noindent Gap~1 ($\kappa_{\mathrm{HH}} - \kappa_{\mathrm{MM}} = 0.110$) is moderate, reflecting probe initialization noise. Gap~2 ($\kappa_{\mathrm{MM}} - \kappa_{\mathrm{MH}} = 0.312$) is $2.8{\times}$ larger: models are self-consistent but converge to a different solution than humans. Since $\kappa_{\mathrm{MM}}$ is estimated from $K\!=\!8$ seeds (split 4+4) with ties broken toward negative, Gap~2 is likely a \emph{lower bound} on true model--human divergence.

\begin{figure}[!t]
\centering
\includegraphics[width=\linewidth]{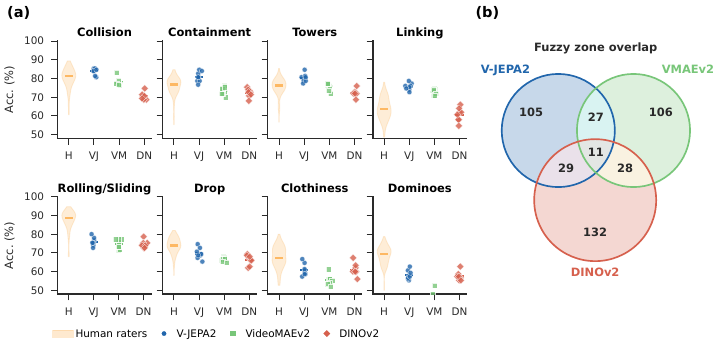}
\caption{\textbf{Models are consistent but differently consistent than humans.} (a)~Per-scenario variability of human raters (gray violins, ${\sim}100$ raters) versus model seeds (colored dots, 8 seeds per model). Human rater distributions are $2$--$3{\times}$ wider ($\sigma\!=\!5$--$8\%$) than model seed distributions ($\sigma\!=\!1$--$3\%$), yet model clusters are \emph{offset} from the human median on most scenarios. (b)~Fuzzy zone overlap: each model's ``fuzzy zone'' (near-chance internal agreement) contains $14$--$17\%$ of samples, but pairwise Jaccard overlap is only ${\sim}0.12$, meaning models are uncertain about different samples.}
\label{fig:rater_vs_seed}
\end{figure}

\subsection{Distributional Divergence}
\label{sec:distributional}

\noindent\textbf{Finding 2:} \emph{Model populations are $2$--$3{\times}$ more internally consistent than humans ($\sigma\!=\!1$--$3\%$ vs.\ $5$--$8\%$), yet converge to systematically different solutions.}

\paragraph{Variability comparison.}
Figure~\ref{fig:rater_vs_seed} visualizes the core finding. Model seeds cluster tightly ($\sigma\!=\!1$--$3\%$) compared to human raters ($\sigma\!=\!5$--$8\%$), but the clusters are systematically \emph{offset} from the human median on most scenarios. Inter-seed agreement is $84.2\%{\pm}1.0\%$, meaning ${\sim}16\%$ of samples yield different predictions across probe initializations; yet the model--human agreement rate ($73.6\%$) is substantially lower, confirming that the model--human gap exceeds internal variability.

\paragraph{Fuzzy zone analysis.}
Each model places $14$--$17\%$ of samples in its ``fuzzy zone'' (3--5 of 8 seeds correct), but the pairwise Jaccard overlap between fuzzy zones is only ${\sim}0.12$: different models are uncertain about largely \emph{different} samples (Figure~\ref{fig:rater_vs_seed}b).

\subsection{Structured Divergence}
\label{sec:structured}

\noindent\textbf{Finding 3:} \emph{Divergence is scenario-structured: models excel on geometry/linking, while human advantages appear in several released-split future-dynamics scenarios, and this structure is captured by strategy fingerprints.}

\begin{table}[t]
\centering
\caption{Per-scenario accuracy (mean${\pm}$std across seeds) for three models and humans. $\Delta$ = \vjepa{} $-$ Human. Best model per scenario in \textbf{bold}.}
\label{tab:scenario}
\small\setlength{\tabcolsep}{3pt}
\begin{tabular}{lcccccc}
\toprule
Scenario & \vjepa{} & \vmae{} & \dino{} & Human & $\Delta$ \\
\midrule
Collision    & $\mathbf{83.8{\pm}1.7}$ & $77.8{\pm}2.2$ & $70.6{\pm}1.8$ & 80.3 & $+3.5$ \\
Containment  & $\mathbf{80.8{\pm}2.6}$ & $73.4{\pm}1.9$ & $72.2{\pm}2.1$ & 76.0 & $+4.8$ \\
Towers       & $\mathbf{80.3{\pm}2.1}$ & $75.8{\pm}1.9$ & $72.2{\pm}1.9$ & 75.5 & $+4.8$ \\
Linking      & $\mathbf{75.6{\pm}1.8}$ & $72.7{\pm}1.4$ & $60.7{\pm}3.4$ & 63.8 & $+11.8$ \\
Roll/Slide   & $75.9{\pm}2.1$ & $\mathbf{74.8{\pm}2.4}$ & $74.6{\pm}1.8$ & 87.7 & $-11.8$ \\
Drop         & $\mathbf{69.7{\pm}2.8}$ & $66.4{\pm}1.0$ & $66.2{\pm}2.5$ & 74.0 & $-4.3$ \\
Clothiness   & $60.9{\pm}3.0$ & $55.5{\pm}2.6$ & $\mathbf{61.1{\pm}3.1}$ & 67.7 & $-6.8$ \\
Dominoes     & $\mathbf{58.2{\pm}2.3}$ & $44.8{\pm}3.6$ & $57.5{\pm}2.2$ & 68.7 & $-10.5$ \\
\midrule
\textbf{Overall} & $\mathbf{73.2{\pm}0.95}$ & $67.7{\pm}0.72$ & $66.9{\pm}0.73$ & 74.2 & $-1.0$ \\
\bottomrule
\end{tabular}
\end{table}

\paragraph{Scenario-level patterns.}
Table~\ref{tab:scenario} reveals consistent patterns across architectures. \vjepa{} outperforms humans on collision, containment, towers, and linking (spatial/geometric scenarios); all three models underperform on \emph{rolling/sliding}, \emph{dominoes}, and \emph{clothiness} in the released split. The largest model advantage is \vjepa{}'s $+11.8$\,pp on linking; the largest released-split human advantage is $-11.8$\,pp on rolling/sliding. Because a subset of rolling/sliding ledge trials is rendering-sensitive, we treat this gap as descriptive of the reviewed benchmark split rather than as standalone evidence for a specific gravitational-dynamics mechanism.

Notably, scenario rankings \emph{partially} differ across models: \dino{} (image-only) matches \vjepa{} on clothiness ($61.1\%$ vs.\ $60.9\%$) and dominoes ($57.5\%$ vs.\ $58.2\%$), suggesting these scenarios rely more on spatial than temporal cues. But \dino{} collapses on linking ($60.7\%$, $-$14.9pp vs.\ \vjepa{}), where temporal chain dynamics are critical. At the 45-subtask level, rank reversals occur across models (\eg on Dominoes 4-Mid, \dino{} reaches $95\%$ while \vmae{} reaches $43\%$), with gaps up to 52\,pp on the same subtask, confirming qualitatively different strategies. To make these aggregate reversals inspectable at the stimulus level, the supplementary material adds representative high-confidence disagreement cases selected from real Physion trials. These cases cover both directions of model--human reversal: human-correct/model-wrong samples whose correct label depends on unobserved future dynamics, and model-correct/human-wrong samples where humans are misled by plausible but incorrect trajectories. The qualitative panels show that divergence is not merely random label noise; they give concrete examples of disagreements around future contact, causal chains, and ambiguous trajectories.

\begin{figure}[!t]
\centering
\includegraphics[width=\linewidth]{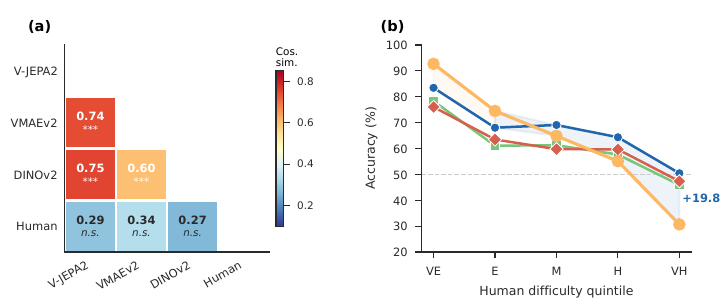}
\caption{\textbf{Strategy and difficulty divergence.} (a)~Strategy fingerprint similarity matrix: cosine similarity between Ridge-regression strategy vectors (37 physical features $\to$ per-sample accuracy). All three models form a tight cluster (cosine$\,{=}\,0.60$--$0.75$, $p\!<\!0.001$), but \emph{no model's strategy is significantly similar to humans} ($p\!>\!0.1$ for all). (b)~Difficulty alignment reversal: model accuracy declines less steeply than human accuracy as stimuli get harder. On the hardest human-defined quintile, \vjepa{} scores $+19.8$\,pp higher while remaining near chance ($r\!=\!0.38$).}
\label{fig:fingerprint}
\end{figure}

\paragraph{Strategy fingerprinting.}
Using the 37-feature core subset defined in \S\ref{sec:fingerprint_method}, Figure~\ref{fig:fingerprint}a shows the pairwise cosine similarity between strategy fingerprints.
Two patterns emerge:
(i)~all model pairs are significantly similar (cosine $0.60$--$0.75$, all $p\!<\!0.001$), forming a tight cluster of shared non-human strategies;
(ii)~\emph{no model aligns with humans} (cosine $0.27$--$0.34$, all $p\!>\!0.1$).
The resulting behavioral fingerprints form a shared non-human cluster. The top model coefficients (\texttt{cutoff\_target\_moving}, \texttt{speed\_trajectory\_var}) relate to dynamics at the observation cutoff, while human correctness is more strongly associated with scene-level properties (\texttt{n\_distractors}, \texttt{total\_kinetic\_energy\_max}) that receive little weight in the model fingerprints.

\paragraph{Future outcomes are associated with divergence.}
For the 3.2\% of stimuli where the target already contacts the zone at the observation cutoff, model--human disagreement is just $10.5\%$ (baseline: no forward reasoning required); for the remaining stimuli requiring future prediction, disagreement rises to $26.8\%$ ($26.8/10.5 = 2.6{\times}$, $p\!<\!0.05$). Crucially, prediction horizon distance does not predict disagreement ($r\!=\!{-}0.065$, $p\!=\!0.12$): clothiness has the shortest horizon yet the highest disagreement, suggesting that divergence depends more on \emph{which type} of reasoning is required than on how far forward the prediction extends.

\subsection{Difficulty Alignment and Error Structure}
\label{sec:complementarity_main}

\noindent\textbf{Finding 4:} \emph{Models and humans find different stimuli difficult: model--human difficulty correlation is moderate ($r\!=\!0.38$), and on the hardest stimuli for humans, models score $+19.8$\,pp higher.}

\paragraph{Difficulty alignment.}
On ``very hard'' samples (human accuracy $<$50\%), humans score $30.7\%$ while \vjepa{} scores $50.4\%$ ($+19.8$\,pp). This apparent model advantage is a useful diagnostic precisely because it is counterintuitive: the bin is selected by human difficulty, not by physical simplicity or model performance. Therefore, the gap should be read as difficulty misalignment rather than stronger physical reasoning: \vjepa{} is near chance on these samples, while the entropy analysis shows that model confidence is often high where humans are uncertain. The model--human difficulty correlation is moderate ($r\!=\!0.38$), leaving ${\sim}86\%$ of model difficulty variance unexplained by human difficulty.

\paragraph{Error overlap structure.}
Pairwise agreement between model majority votes is only ${\sim}$72--77\%, substantially less than human--human agreement ($95.2\%$).
Error consistency analysis (Geirhos~\etal~\cite{GeirhosError2020}) reveals that model--model error overlap (avg.\
Jaccard$\,{=}\,0.41$) far exceeds model--human overlap (avg.\ Jaccard$\,{=}\,0.26$): models share error patterns with each other but not with humans.

\paragraph{Robustness on Physion++.}
We replicate the analysis on Physion++~\cite{Tung2024}, which tests latent physical properties (mass, friction, elasticity, deformability) across 10 scenarios. This robustness check asks whether the disagreement pattern is tied only to the original Physion item distribution or persists when the benchmark stresses different physical properties. Under the matched supplementary protocol, \vjepa{} reaches $70.44\%$ accuracy on Physion++, while model--human disagreement rises from $27.0\%$ on Physion (cf.\ $26.4\%$ under the main-text protocol) to $50.1\%$ on Physion++; the deformability condition also reverses the apparent ranking between model and human performance. Because the protocol and stimuli are not identical to the main benchmark, we report Physion++ as supplementary evidence rather than a replacement for the main analysis, but it supports the conclusion that the divergence is not a narrow artifact of the original Physion split (details in supplementary material).

\section{What Drives Divergence?}
\label{sec:diagnostic}

The distributional analysis in \S\ref{sec:results} establishes \emph{that} models and humans diverge. We now ask which classes of physical features are associated with this divergence, testing whether the accessible signal lies primarily in visible information or in unobserved future outcomes.

\paragraph{Feature attribution.}
Using the full 89-feature set defined in \S\ref{sec:fingerprint_method}, which expands the 37-feature strategy fingerprinting subset to enable layer-wise attribution, we separate features into \textbf{A}~(19 static/scene), \textbf{B}~(47 observable dynamics), and \textbf{C}~(23 unobservable outcome). This analysis directly tests whether model--human divergence can be explained by visible scene information available before the prediction cutoff. Across Ridge regression, random forest, and gradient boosting models evaluated by 5-fold cross-validation, the best observable-feature model (A+B, 66 dimensions) still yields \emph{negative} $R^2\!=\!{-}0.32$ for predicting model--human divergence $|p_m - p_h|$ (AUROC $=0.63$). We cannot rule out other nonlinear relationships, but this suggests observable features carry little accessible information about divergence under this protocol. Unobservable outcome features (C, 23 dimensions) achieve $R^2\!=\!0.29$ and AUROC$\,{=}\,0.70$; for ``both-certain disagree'' samples, AUROC reaches $0.85$. This asymmetry is consistent with divergence associated with forward outcomes rather than scene encoding alone. We note that Layer~C features are correlated with task outcome by construction, so this attribution is suggestive rather than conclusive; disentangling the two would require interventional experiments beyond the scope of this work. The supplementary material reports all regressors and targets to make the negative observable-feature result and the positive unobservable-feature result transparent.

\paragraph{Calibration and confidence.}
Entropy decomposition reveals a notable asymmetry ($A = |Q2|/|Q3| = 0.31$--$0.39$ across models; $0.32$ for \vjepa{}): models are more often confident where humans are uncertain (Q3) than uncertain where humans are certain (Q2). This does not imply superior reasoning: models confidently commit to answers on samples where humans remain split, consistent with a qualitatively different strategy rather than additional knowledge. However, within disagreement samples, overconfidence is prevalent: \vjepa{} is the best-calibrated (ECE$\,{=}\,0.166$, only 3 high-confidence errors vs.\ 16--24 for alternatives), but all models assign confident predictions to samples they get wrong. Full calibration and surprise analysis details are in supplementary material.

\section{Discussion}
\label{sec:discussion}

\paragraph{Accuracy as an Incomplete Metric.}
Our central finding is that near-identical accuracy ($73.2\%$ vs.\ $74.2\%$) coexists with systematic behavioral divergence structured by scenario and future-prediction demands. The three-level agreement hierarchy quantifies this: Gap~1 ($\kappa_{\mathrm{HH}} - \kappa_{\mathrm{MM}} = 0.110$) is moderate, but Gap~2 ($\kappa_{\mathrm{MM}} - \kappa_{\mathrm{MH}} = 0.312$) is $2.8{\times}$ larger, representing irreducible divergence between two populations that converge to \emph{different} solutions.

\paragraph{Why Do Models Converge to Different Solutions?}
The strategy fingerprint analysis is consistent with \emph{information access asymmetry} at the behavioral level: human correctness patterns reflect scenario-specific priors over object dynamics and causal chains that generalize from real-world experience, whereas the evaluated frozen representations show a more shared cue-based fingerprint that succeeds where geometry is sufficient but struggles where future dynamics are critical. We cannot attribute this pattern uniquely to objective, architecture, training data, or the supervised probe protocol; a supplementary recurrent-model check supports the same signature within this protocol. The Jaccard overlap of ${\sim}0.12$ between model fuzzy zones suggests architecture-specific blind spots within this broader pattern. The difficulty alignment reversal should also be read as difficulty misalignment rather than model superiority: on stimuli hardest for humans ($<$50\% accuracy), \vjepa{} is near chance but still $+19.8$\,pp above the human bin average, implying that the factors making stimuli hard differ between populations.

\paragraph{Implications for Benchmark Design.}
Our framework reveals two limitations of current practice. First, \emph{reporting accuracy alone is insufficient}: we advocate including $\kappa_{\mathrm{MH}}$ and disagreement rate as standard metrics. Second, \emph{multi-seed evaluation should be standard}: the ${\sim}15\%$ inter-seed disagreement rate means any single-seed result is a noisy estimate, and population-level comparison requires multiple runs. The ${\sim}5$--$6$pp gap between \vjepa{} and alternatives only became apparent with unified ViT-L backbones.

\paragraph{Confident Divergence, Not Superiority.}
The entropy asymmetry ($A\!=\!0.31$--$0.39$ across models; \S\ref{sec:diagnostic}) shows that Q3 (model certain, humans uncertain) outnumbers Q2 by about $3{\times}$, but model accuracy on Q3 samples is not above chance for the hardest cases. Models do not ``know more''; they commit confidently to a different answer. This is consistent with a cue-based strategy that can produce confident outputs even on genuinely ambiguous stimuli, without matching human uncertainty.

\paragraph{Scene-Level Cues vs.\ Forward Simulation.}
Our findings are consistent with a dual-process account~\cite{Battaglia2013,Kubricht2017}: models excel on geometry-driven scenarios (linking $+11.8$\,pp), whereas human advantages appear on causal-chain judgments, including dominoes ($-10.5$\,pp). We report rolling/sliding ($-11.8$\,pp) as a released-split descriptive gap, not a standalone mechanism diagnostic, because ledge trials are rendering-sensitive. The cutoff-contact analysis (\S\ref{sec:structured}) suggests that the relevant distinction is not temporal extrapolation distance alone, but whether the future dynamics needed for the decision are accessible to the frozen representation and probe. We emphasize that our analysis provides behavioral and attributional constraints on these accounts, not direct evidence of internal mechanisms; forward-simulation-like representations may exist within models but remain untapped by this evaluation protocol.

\paragraph{Broader Applicability.}
Our distributional framework applies wherever per-sample human judgments are available alongside multiple model runs. The agreement hierarchy, entropy decomposition, and strategy fingerprinting are general tools that could reveal similar convergence illusions in other domains.

\paragraph{Limitations.}
Our framework requires multiple human raters per stimulus, so it applies most directly to crowd-sourced benchmarks. The main analysis uses one binary contact benchmark; Physion++ provides robustness evidence but not a complete benchmark sweep. Some released-split scenario subtypes are item-quality sensitive, so no single scenario gap is a definitive mechanism diagnostic. We evaluate ViT-L frozen representations with supervised probes, so larger 1B+ models, fine-tuned systems, generative/violation-of-expectation protocols, or interactive settings may yield different divergence patterns. Training data and benchmark familiarity are potential confounds that we do not causally isolate. Finally, strategy fingerprints are correlational summaries based on the 37-feature core subset (89-feature set for attribution); richer feature sets or interventional experiments may reveal additional structure.

\section{Conclusion}
\label{sec:conclusion}

Our analysis yields a simple diagnostic: when a model matches human accuracy on a physical-reasoning benchmark, the next question should not be ``has the gap closed?'' but ``do model and human errors cluster in the same places?'' On Physion, the answer in our setting is no. The $\kappa_{\mathrm{MH}}$ of $0.48$, compared to $\kappa_{\mathrm{HH}}$ of $0.91$, quantifies a behavioral gap that persists across three ViT-L architectures and is structured by the physical demands of each scenario: models excel where geometric cues suffice, while human advantages appear in several future-dynamics and causal-chain scenarios.

The locus of this gap, as indicated by feature attribution, lies downstream of scene encoding: observable scene properties carry little accessible predictive signal for where models and humans diverge within the scope of this evaluation ($R^2\!=\!{-}0.32$), whereas unobservable outcome features are more predictive ($R^2\!=\!0.29$). These results do not show that models lack all relevant physical information; rather, they show that human-level accuracy can coexist with a different treatment of unobserved outcomes, a pattern shared across the evaluated architectures and invisible to accuracy-based evaluation.

These findings motivate two concrete next steps: applying distributional evaluation wherever per-sample human judgments are available, and testing interactive or generative forward-simulation protocols that separate future-dynamics errors from task-understanding or perceptual-ambiguity effects.

\section*{Acknowledgements}
We thank the creators of the Physion benchmark for making their data and human response data publicly available.

\bibliographystyle{splncs04}
\bibliography{main}

\newif\ifincludeappendix
\includeappendixfalse  
\ifincludeappendix
\appendix

\section{Detailed Cross-Model Results Including RVM}
\label{sec:rvm_extended}

As a supplementary robustness check, we evaluate \rvm{}~\cite{zoran2025}, a recurrent vision model that processes frames sequentially via a GRU-augmented ViT-L backbone.
\rvm{} introduces a qualitatively different inductive bias (recurrence) while satisfying the same fair-comparison constraints (ViT-L backbone, E2E attentive probes, multi-seed evaluation).
If the model--human divergence we report is specific to the three main-text architectures, \rvm{} should break the pattern; if it reflects a broader pattern under our frozen-representation/probe protocol, \rvm{} should replicate it.
Tables~\ref{tab:crossmodel_ext} and~\ref{tab:scenario_ext} provide the detailed \rvm{} cross-model and per-scenario results.

\paragraph{Key finding.}
\rvm{} ($73.1{\pm}1.1\%$, 10 seeds) performs comparably to \vjepa{} ($73.3{\pm}0.9\%$ in the same supplementary 10-seed protocol); the difference is not statistically significant ($p{=}0.68$, Welch's $t$, Cohen's $d{=}0.19$).
Its $\kappa_{\mathrm{MH}}{=}0.491$ is slightly higher than \vjepa{}'s ($0.464$), and its disagreement rate ($25.2\%$) is lower than \vjepa{}'s ($26.9\%$) despite slightly lower accuracy---reflecting that \rvm{}'s errors overlap more with human errors (i.e., it errs on samples humans also find difficult). Both models remain far below $\kappa_{\mathrm{HH}}{=}0.906$.
\rvm{}'s strategy fingerprint is most similar to \vjepa{} (cos$\,{=}\,0.83$, $p{<}0.001$) and least similar to humans (cos$\,{=}\,0.18$, $p{=}0.39$), suggesting that the non-human behavioral signature identified in the main text is not an artifact of the three non-recurrent architectures examined in the selected-model panels, but also appears for a fourth, architecturally distinct model under the same protocol.

\begin{table}[!htbp]
\centering
\caption{Supplementary 10-seed cross-model comparison including \rvm{} (all ViT-L, unified protocol).}
\label{tab:crossmodel_ext}
\small\setlength{\tabcolsep}{4pt}
\begin{tabular}{lccccc}
\toprule
Model & Seeds & Test Acc. & Disagr. & $\kappa_{\mathrm{MH}}$ \\
\midrule
\vjepa{}      & 10 & $\mathbf{73.3{\pm}0.9}$ & $26.9$ & $0.464$ \\
\rvm{}        & 10 & $73.1{\pm}1.1$ & $25.2$ & $0.491$ \\
\vmae{}       & 10  & $67.8{\pm}0.8$ & $30.7$ & $0.385$ \\
\dino{}       & 10  & $66.9{\pm}0.7$ & $32.0$ & $0.364$ \\
Human\textsuperscript{\dag}         & ${\sim}$100 & $74.2$ & $4.8$ & $0.906$ (HH) \\
\bottomrule
\end{tabular}
\end{table}

\begin{table}[!htbp]
\centering
\caption{Extended per-scenario accuracy including \rvm{}. $\Delta$ = \vjepa{} $-$ Human.}
\label{tab:scenario_ext}
\small\setlength{\tabcolsep}{2.5pt}
\begin{tabular}{lcccccc}
\toprule
Scenario & \vjepa{} & \rvm{} & \vmae{} & \dino{} & Human & $\Delta$ \\
\midrule
Collision    & $\mathbf{83.8{\pm}1.6}$ & $83.7{\pm}1.5$ & $78.5{\pm}2.0$ & $70.1{\pm}2.0$ & 80.3 & $+3.5$ \\
Containment  & $\mathbf{81.4{\pm}2.6}$ & $79.0{\pm}1.9$ & $73.8{\pm}1.9$ & $72.7{\pm}2.3$ & 76.0 & $+5.4$ \\
Towers       & $\mathbf{80.5{\pm}2.4}$ & $79.1{\pm}2.1$ & $76.1{\pm}2.0$ & $72.0{\pm}2.2$ & 75.5 & $+5.0$ \\
Linking      & $\mathbf{75.8{\pm}1.6}$ & $74.7{\pm}1.7$ & $72.9{\pm}1.5$ & $60.9{\pm}3.3$ & 63.8 & $+12.0$ \\
Roll/Slide   & $75.6{\pm}3.3$ & $\mathbf{75.5{\pm}2.0}$ & $74.7{\pm}3.5$ & $74.5{\pm}3.3$ & 87.7 & $-12.1$ \\
Drop         & $\mathbf{68.9{\pm}2.9}$ & $67.7{\pm}3.2$ & $66.9{\pm}1.3$ & $66.5{\pm}2.4$ & 74.0 & $-5.1$ \\
Clothiness   & $61.6{\pm}3.2$ & $\mathbf{66.9{\pm}2.6}$ & $55.3{\pm}2.8$ & $60.8{\pm}2.8$ & 67.7 & $-6.1$ \\
Dominoes     & $\mathbf{58.5{\pm}2.2}$ & $57.9{\pm}3.3$ & $44.5{\pm}3.4$ & $58.1{\pm}2.5$ & 68.7 & $-10.2$ \\
\midrule
\textbf{Overall} & $\mathbf{73.3{\pm}0.9}$ & $73.1{\pm}1.1$ & $67.8{\pm}0.8$ & $66.9{\pm}0.7$ & 74.2 & $-0.9$ \\
\bottomrule
\end{tabular}
\end{table}

\begin{figure}[!htbp]
\centering
\includegraphics[width=\linewidth]{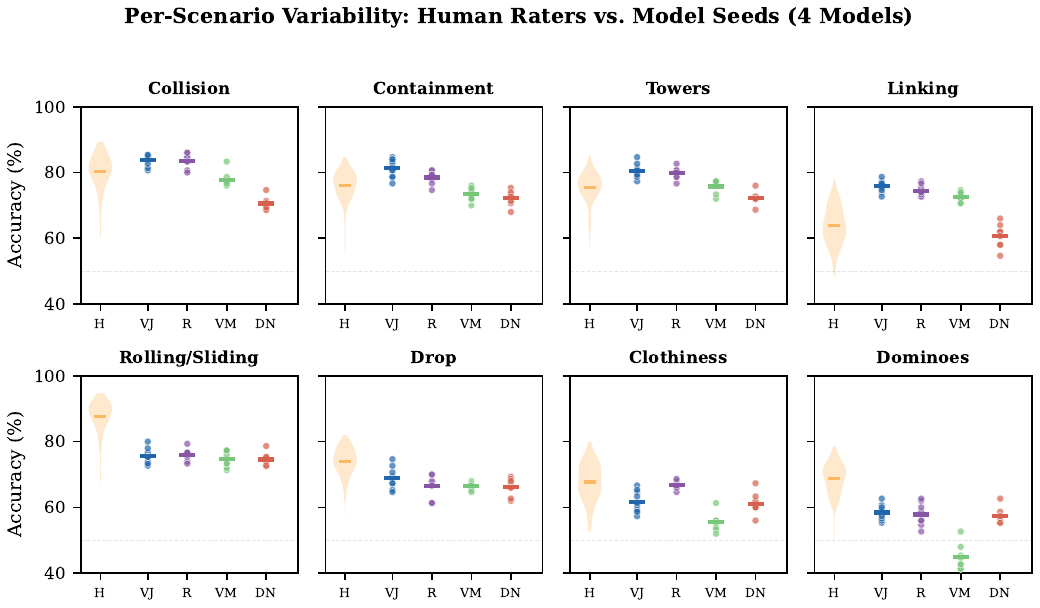}
\caption{\textbf{Extended variability comparison (4 models).} Per-scenario variability of human raters (amber violins) versus model seeds (colored dots). \rvm{} (purple) clusters as tightly as \vjepa{} (blue) and tracks the same per-scenario pattern, with both offset from the human median.}
\label{fig:ext_rater_vs_seed}
\end{figure}

\begin{figure}[!htbp]
\centering
\includegraphics[width=\linewidth]{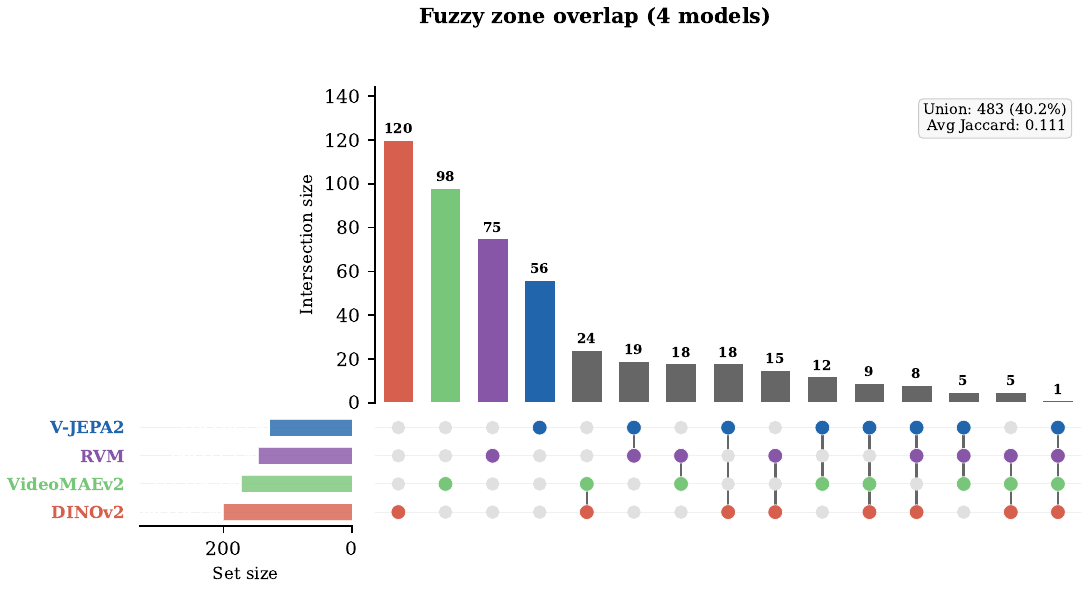}
\caption{\textbf{Fuzzy zone overlap (4 models, UpSet plot).} Left: set sizes (each model's fuzzy zone count). Top: intersection sizes sorted by count. Dot matrix below indicates set membership. The four exclusive (single-model) regions dominate ($56$--$120$ samples each), while pairwise overlaps are small ($\leq 24$). Only 1 sample lies in the intersection of all four fuzzy zones, confirming that models are uncertain about largely different samples (avg.\ Jaccard$\,{=}\,0.111$).}
\label{fig:ext_fuzzy_zone}
\end{figure}

\begin{figure}[!htbp]
\centering
\includegraphics[width=0.6\linewidth]{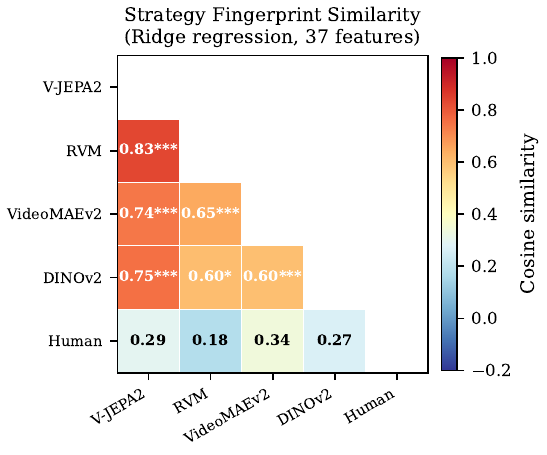}
\caption{\textbf{Extended strategy fingerprint (4 models + human).} Cosine similarity between Ridge-regression strategy vectors. \rvm{} is most similar to \vjepa{} ($0.83$, $p\!<\!0.001$) and least similar to humans ($0.18$, $p\!=\!0.39$). All four models form a tight cluster; no model signature aligns with the human signature.}
\label{fig:ext_fingerprint}
\end{figure}

\begin{figure}[!htbp]
\centering
\includegraphics[width=0.65\linewidth]{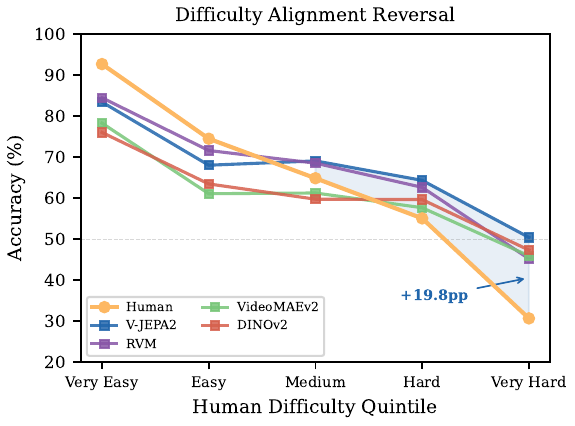}
\caption{\textbf{Extended difficulty alignment (4 models).} On the hardest quintile, all four models outperform humans. \rvm{} shows a pattern nearly identical to \vjepa{}, reinforcing that these architecturally different models converge on the same difficulty profile.}
\label{fig:ext_difficulty}
\end{figure}

\clearpage

\section{Layer-wise Probing Analysis}
\label{sec:layerwise_appendix}

\noindent\emph{Note: The layer-wise probing experiments in this section were conducted during an earlier experimental phase using frame\_step$\,{=}\,$5 (1.17\,s coverage). The main-text analyses and supplementary cross-model comparison in \S\ref{sec:rvm_extended} use the final protocol with frame\_step$\,{=}\,$5.75 (1.50\,s, matching the human observation window). Because this section examines relative layer-wise trends within a single model rather than absolute accuracy or model--human comparison, the qualitative findings (peak at layer~16, plateau at L15--L22, final-layer drop) are robust to this difference in temporal coverage.}

Physical reasoning accuracy peaks at layer~16 of 24 ($71.83\%$) within a broad plateau (layers 15--22), dropping $-$3.0pp at the final layer.
This is consistent with the ``Physics Emergence Zone'' of concurrent work~\cite{Joseph2026}.
Linear probes peak later (layer~19) and are ${\sim}10$pp lower, suggesting that stronger probes extract physics information from earlier layers.

\paragraph{Prediction trajectories.}
Only $19.9\%$ of samples are always correct across all layers; $24.2\%$ are correctly classified at intermediate layers but lost at the final layer (``peak then lost''), concentrated in dominoes ($34.7\%$) and, descriptively in the released split, rolling/sliding ($27.3\%$). Because this earlier analysis uses the released split, we do not treat the rolling/sliding concentration alone as clean mechanism evidence; the layer-wise trend is most useful as a within-protocol diagnostic of information loss in deeper layers.

\begin{figure}[!htbp]
\centering
\includegraphics[width=\linewidth]{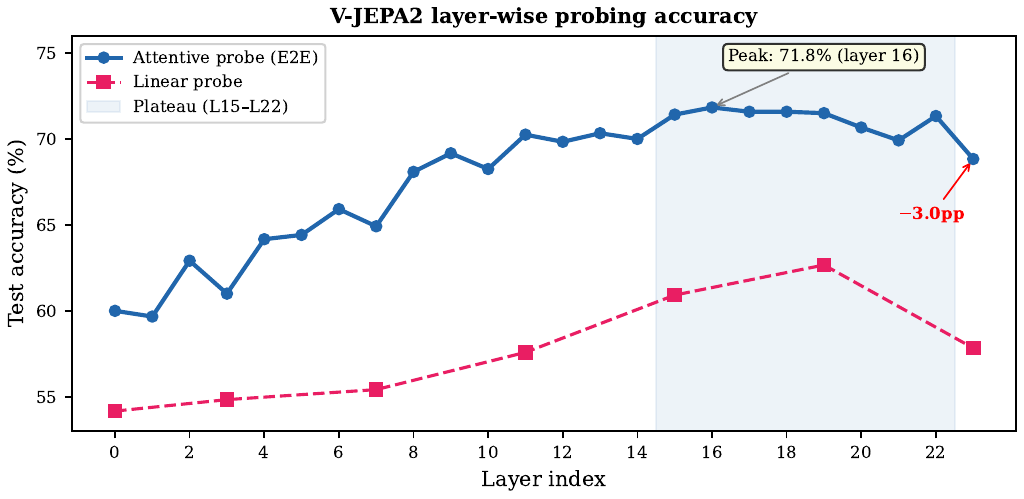}
\caption{Layer-wise probing accuracy for \vjepa{} (ViT-L, 24 layers). Attentive probes (blue) peak at layer~16 ($71.8\%$) within a broad plateau (L15--L22, shaded), then drop $-$3.0pp at the final layer. Linear probes (pink) peak later (layer~19) and are ${\sim}10$pp lower throughout, consistent with the capacity gap expected from a 2K-parameter linear head.}
\label{fig:layerwise}
\end{figure}

\section{Spatial Token Structure Analysis}
\label{sec:token_appendix}

\noindent\emph{Note: token-level probes use frame\_step$\,{=}\,$5 (1.17\,s coverage), the setting under which the original feature extraction was conducted. Main-text results use frame\_step$\,{=}\,$5.75 (1.50\,s); see \S\ref{sec:layerwise_appendix} header for discussion.}

\begin{table}[!htbp]
\centering
\caption{Token-level probes on \vjepa{} full features [$8{\times}256$, $1024$].}
\label{tab:token_probes}
\begin{tabular}{lcc}
\toprule
Probe & Params & Test Acc. \\
\midrule
\multicolumn{3}{l}{\emph{Mean-pooled baselines}} \\
~~~Linear & 2K & 61.50\% \\
~~~Attentive (depth-4) & 2.4M & 62.50\% \\
\midrule
\multicolumn{3}{l}{\emph{Token-level probes}} \\
~~~GRU-Pool & 3.7M & 66.00\% \\
~~~TemporalAttn & 7.9M & 67.33\% \\
~~~GRU-Attn & 12.1M & \textbf{67.42\%} \\
~~~CrossAttn-GRU & 29.4M & 64.75\% \\
\bottomrule
\end{tabular}
\end{table}

All token-level probes outperform mean-pooled baselines by $+$2.3--$4.9$pp. On rolling/sliding, CrossAttn-GRU (preserving 256 spatial tokens) gains $+$15.3\% over GRU-Pool (spatial mean-pooling). GRU and self-attention are statistically indistinguishable under identical spatial pooling (McNemar $p\!=\!0.86$), pointing to spatial structure as the key factor in this probe-capacity comparison.

\section{Marker Ablation}
\label{sec:ablation_appendix}

\noindent\emph{Note: marker ablation uses frame\_step$\,{=}\,$5 (1.17\,s coverage), the setting under which the original ablation was conducted. The overall accuracy of $71.6\%$ here (vs.\ $73.2\%$ in the main text) reflects this shorter observation window, not a protocol discrepancy.}

\begin{table}[!htbp]
\centering
\caption{Red-yellow marker ablation on \vjepa{}.}
\label{tab:ablation}
\small
\begin{tabular}{lccc}
\toprule
Scenario & With & Without & $\Delta$ \\
\midrule
roll/slide  & 80.0 & \textbf{84.0} & $+$4.0 \\
drop        & 63.3 & 66.0          & $+$2.7 \\
towers      & 76.7 & 74.7          & $-$2.0 \\
linking     & 74.7 & 72.7          & $-$2.0 \\
clothiness  & 61.3 & 60.0          & $-$1.3 \\
dominoes    & 59.3 & 56.0          & $-$3.3 \\
collision   & 78.7 & 71.3          & $-$7.3 \\
contain.    & 78.7 & 68.7          & $-$10.0 \\
\midrule
\textbf{Overall} & \textbf{71.6} & \textbf{69.2} & $\mathbf{-2.4}$ \\
\bottomrule
\end{tabular}
\end{table}

The overall $-$2.4pp drop when markers are removed suggests that models use scene-level visual cues, including object identity markers, as part of their prediction pattern.
The pattern is scenario-dependent in an interpretable way: containment shows the largest drop ($-$10.0pp), consistent with the fact that predicting whether an object lands \emph{inside} a container requires identifying which object is the target---a task for which colored markers provide a direct cue.
Conversely, rolling/sliding \emph{improves} by $+$4.0pp without markers in this released-split ablation, suggesting that markers can act as distractors interfering with motion-relevant features (surface geometry, trajectory). We treat this as a protocol-specific cue-use pattern rather than a standalone mechanism claim about rolling/sliding.
This asymmetry further supports the view that models rely on a mixture of scene-level statistical cues rather than a single human-like physical reasoning pattern.

\section{Ruling Out Annotation Noise}
\label{sec:annotation_noise}

\paragraph{Split-half consistency.}
Bootstrap split-half analysis (1,000 iterations, ${\sim}100$ raters per stimulus) yields human--human disagreement of $4.8\%$ ($\kappa_{\mathrm{HH}}\!=\!0.906$). Model--human disagreement ranges from $25.2\%$ (\rvm{}) to $32.0\%$ (\dino{}), $5.3$--$6.7{\times}$ higher. Even on samples where humans show the most internal disagreement (clothiness, $8.9\%$), model--human divergence is $4.0{\times}$ higher ($35.3\%$).

\paragraph{Single-rater comparison.}
\vjepa{}'s $\kappa\!=\!0.469$ against the full human majority vote (single model prediction vs.\ aggregated ground truth) falls below \emph{every} individual human rater (mean single-rater $\kappa\!=\!0.632{\pm}0.021$, 0th percentile).
Note this differs from $\kappa_{\mathrm{MH}}\!=\!0.464$ reported in Table~\ref{tab:crossmodel_ext}, which is computed as per-seed average (each seed vs.\ human majority); the two numbers are methodologically distinct but consistent.

\section{Physion++ Extended Analysis}
\label{sec:physionpp_appendix}

\noindent\emph{Note: The Physion++ experiments were conducted using an earlier protocol with frame step $=$ 5 (1.17\,s temporal coverage) rather than the final frame step $=$ 5.75 used in all main-text Physion analyses. The primary finding of this section---that model--human divergence persists and is amplified on property-inference tasks---concerns relative patterns rather than absolute levels, and is robust to this difference.}

\vjepa{} achieves $70.44\%$ on Physion++ despite 92\% less training data ($-$1.2pp vs.\ the matched Physion baseline). Under the matched frame-step comparison used for this appendix, model--human disagreement rises from $27.0\%$ on Physion (cf.\ $26.4\%$ under the main-text frame-step protocol) to $50.1\%$ on Physion++. For deformability, \vjepa{} achieves $81.2\%$ vs.\ human $45.3\%$ ($+$35.9pp); for friction, \vjepa{} drops to $55.2\%$ vs.\ human $56.8\%$.

\begin{figure}[!htbp]
\centering
\includegraphics[width=\linewidth]{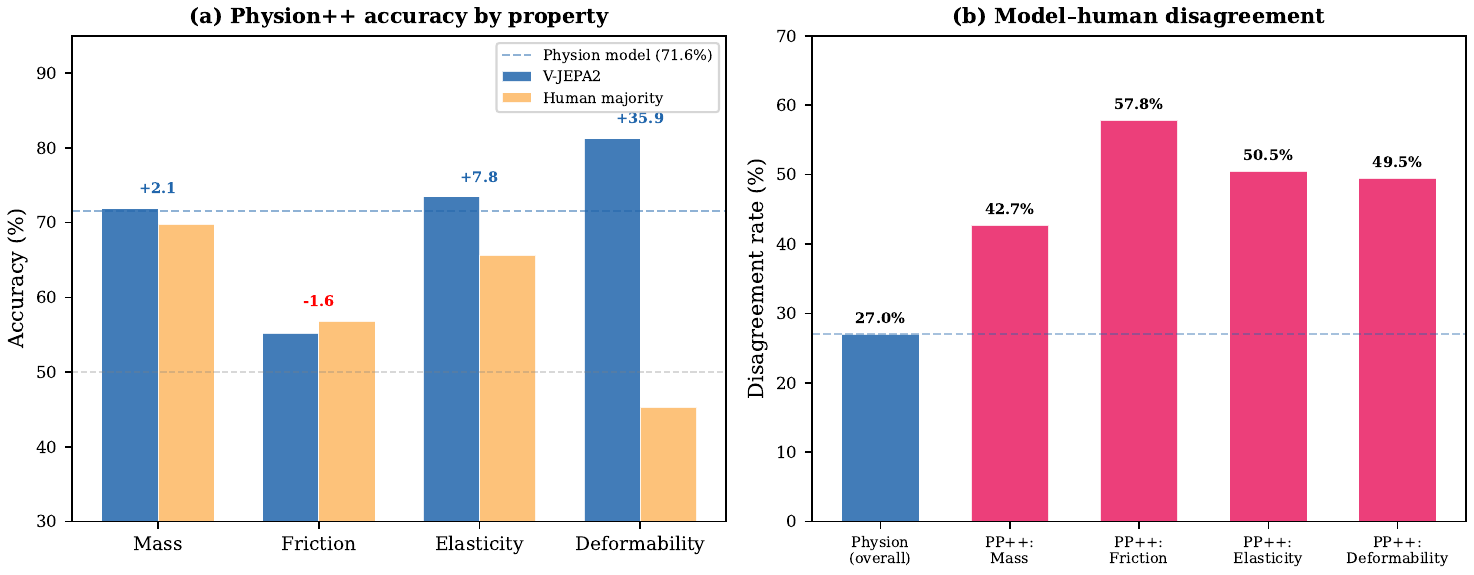}
\caption{\textbf{Physion++ extended analysis.} (a)~Per-property accuracy: \vjepa{} dramatically outperforms humans on deformability ($+$35.9pp) but underperforms on friction ($-$1.6pp). (b)~Model--human disagreement roughly doubles from Physion ($27\%$ at frame\_step$\,{=}\,$5; cf.\ $26.4\%$ at frame\_step$\,{=}\,$5.75 in main text) to Physion++ ($43$--$58\%$), with friction showing the highest disagreement ($57.8\%$).}
\label{fig:physionpp}
\end{figure}

\section{Cross-Model Complementarity}
\label{sec:complementarity_appendix}

\paragraph{Why majority voting fails.}
We test three voting schemes:
all-four majority, top-3 majority, and \vjepa{}+\rvm{} fallback yield $71.8\%$, $72.1\%$, and $72.8\%$, respectively.
Relative to 10-seed \vjepa{} alone, these are $-1.5$, $-1.2$, and $-0.5$pp; \rvm{}'s corrections on \vjepa{} errors are offset by introducing its own.
In all cases, weaker models rescue some errors but overrule more correct predictions (net loss).
\vjepa{}'s ${\sim}5$pp accuracy advantage over \vmae{}/\dino{} makes it consistently outvoted, while \rvm{} ($73.1\%$) is close enough to \vjepa{} ($73.3\%$) that their combined votes do not improve upon \vjepa{} alone.

\paragraph{Unique contributions.}
Each model contributes uniquely correct samples that no other model classifies correctly: \rvm{} contributes 64 unique solves, \dino{} 51, \vjepa{} 46, and \vmae{} 42.
Despite this complementarity, the total pool of uniquely solvable samples (203) is small relative to the shared error mass, explaining why ensemble methods fail to improve over the best individual model.
\dino{}'s unique solves are $73.8\%$ no-contact (consistent with its tendency to predict no-contact more frequently than other models).
\rvm{}, despite contributing the most unique solves, shows the highest pairwise agreement with \vjepa{} ($80.1\%$, $\kappa\!=\!0.605$), suggesting that its recurrent architecture converges on a similar behavioral profile to \vjepa{}'s video transformer.

\section{Per-Subtask Rater vs.\ Seed Variability}
\label{sec:subtask_variability}

For each of the 8 scenarios, we plot per-subtask variability comparing human raters (${\sim}100$ raters, amber bars with $\pm$1$\sigma$ errorbars) and model seeds (colored bars with $\pm$1$\sigma$ errorbars): \vjepa{}, \rvm{}, \vmae{}, and \dino{}.
Key observations:
(i)~Clothiness test12/test19 show high variance for both populations, indicating genuinely uncertain stimuli.
(ii)~Towers nb5 shows the highest human variability ($\sigma\!=\!10.8\%$) with a large model--human gap.
(iii)~Drop sidezone: all models perform far below humans (${\sim}55\%$ vs.\ ${\sim}67\%$).
(iv)~Linking nl4-8: both populations show elevated variability, suggesting high-difficulty chain reasoning.
(v)~\rvm{} generally tracks \vjepa{}'s per-subtask profile but with slightly lower peaks, consistent with its comparable overall accuracy ($73.1\%$ vs.\ $73.3\%$).
Figures~\ref{fig:subtask_collision}--\ref{fig:subtask_dominoes} show all eight scenarios.

\begin{figure}[!htbp]
\centering
\includegraphics[width=\linewidth]{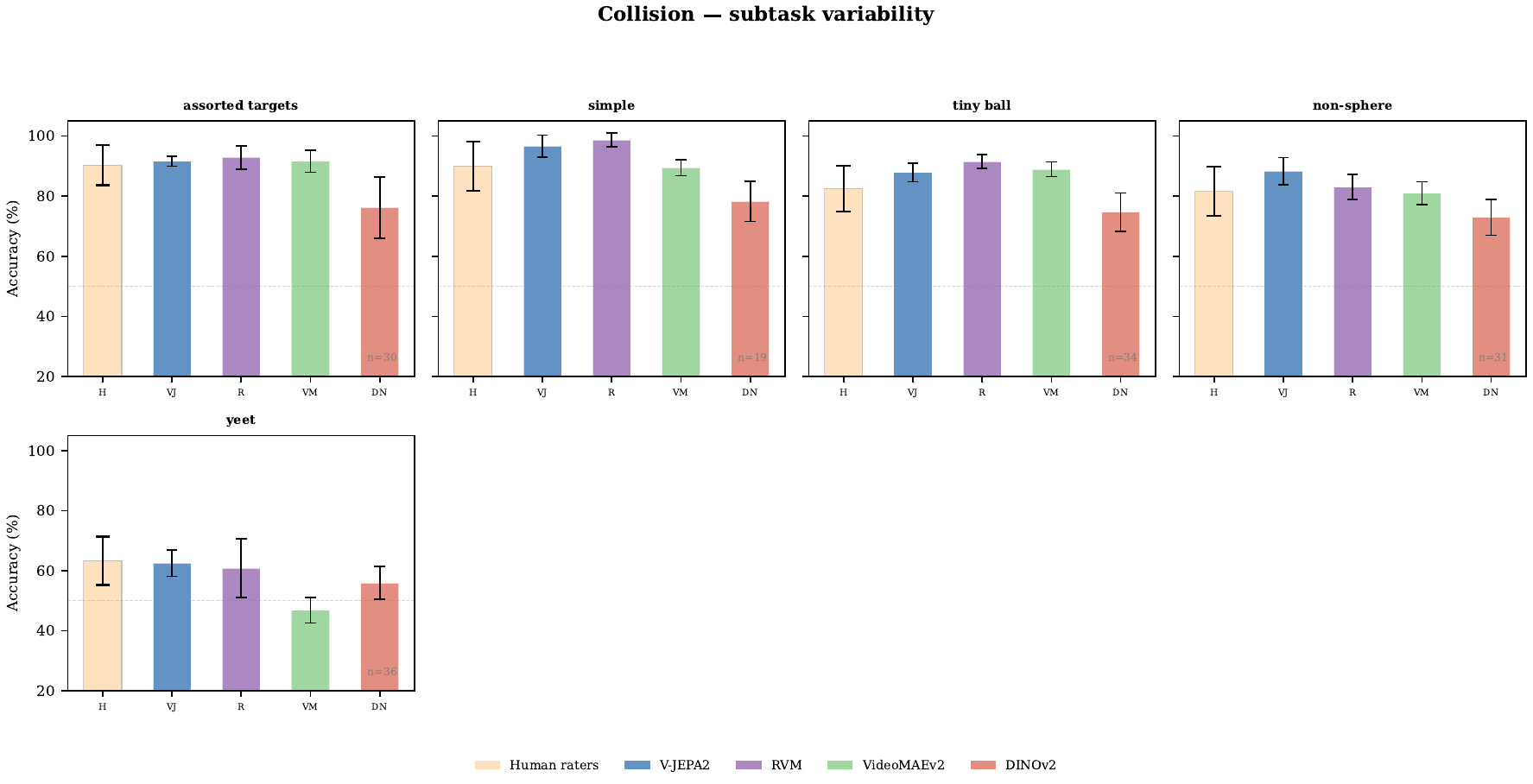}
\caption{Collision subtask variability.}
\label{fig:subtask_collision}
\end{figure}

\begin{figure}[!htbp]
\centering
\includegraphics[width=\linewidth]{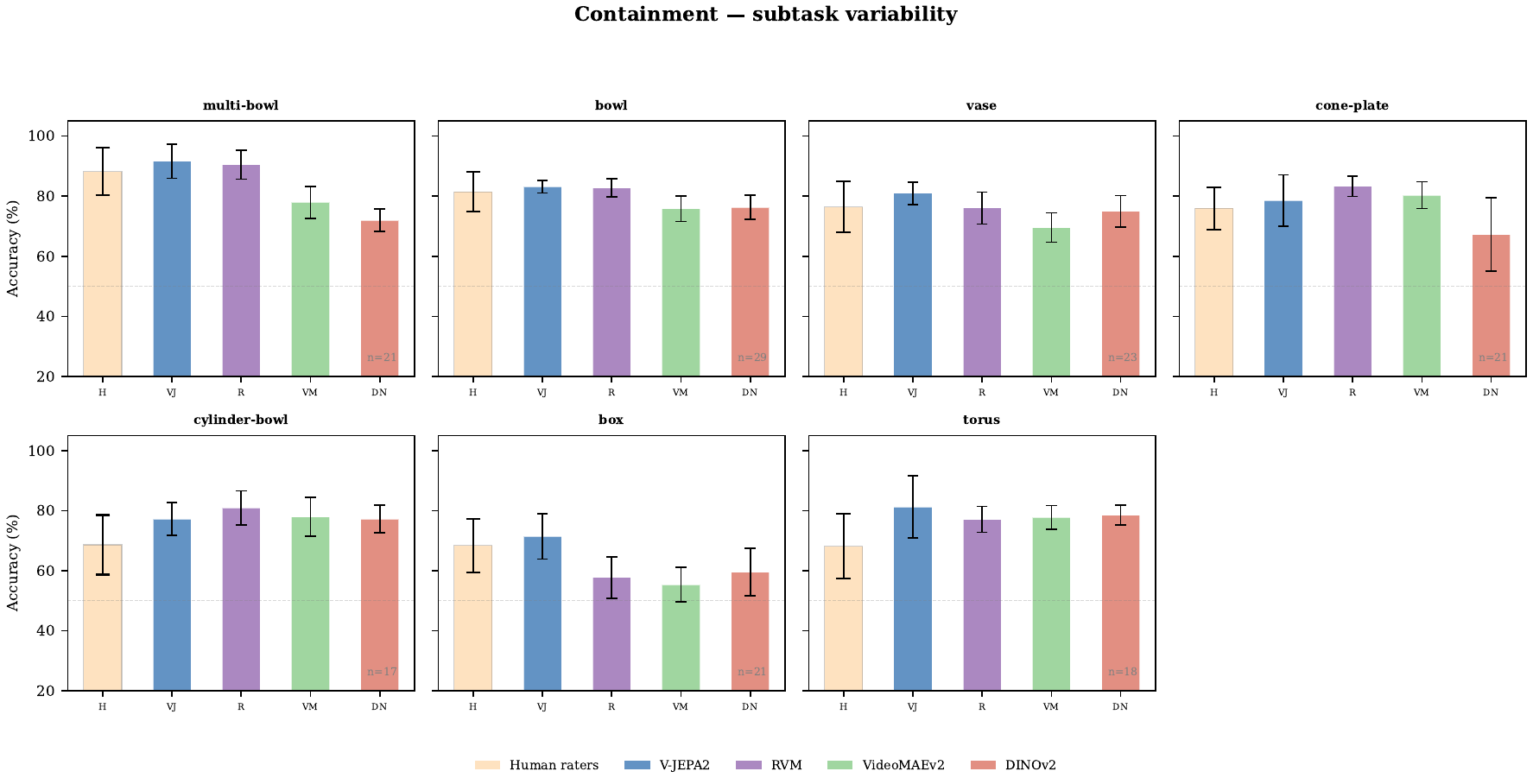}
\caption{Containment subtask variability.}
\label{fig:subtask_containment}
\end{figure}

\begin{figure}[!htbp]
\centering
\includegraphics[width=\linewidth]{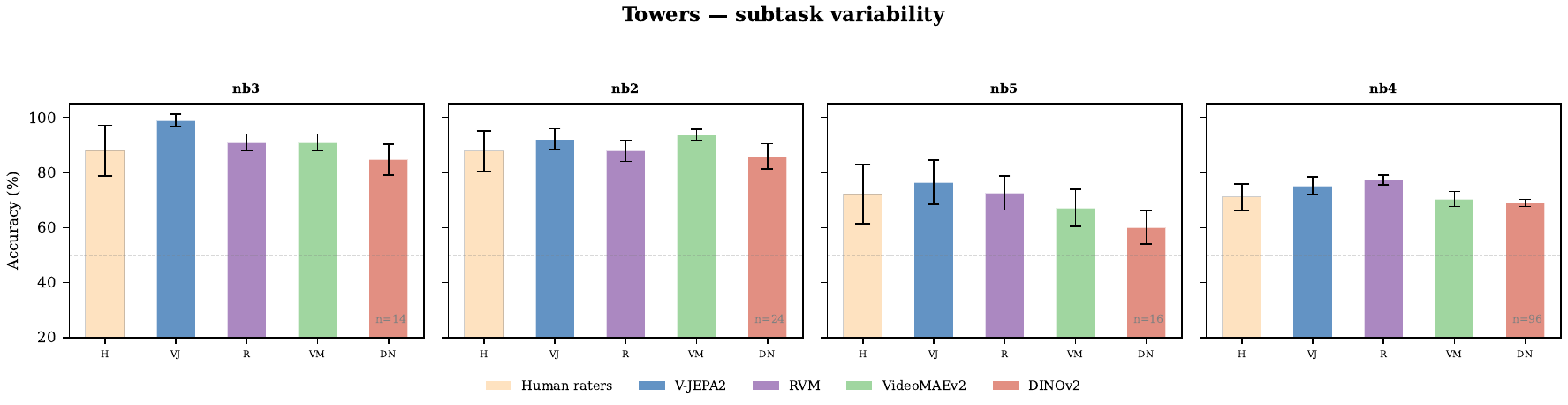}
\caption{Towers subtask variability.}
\label{fig:subtask_towers}
\end{figure}

\begin{figure}[!htbp]
\centering
\includegraphics[width=\linewidth]{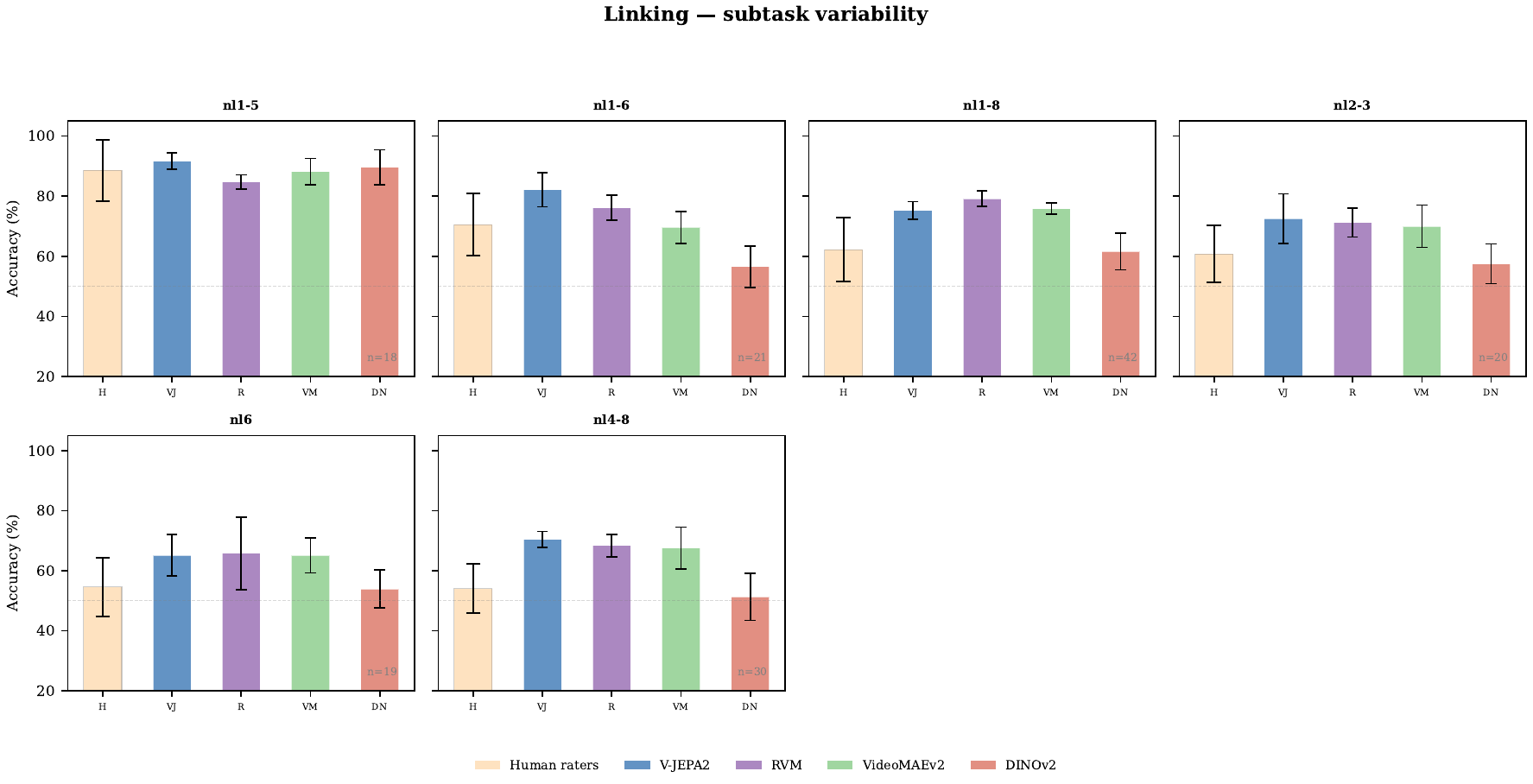}
\caption{Linking subtask variability.}
\label{fig:subtask_linking}
\end{figure}

\begin{figure}[!htbp]
\centering
\includegraphics[width=\linewidth]{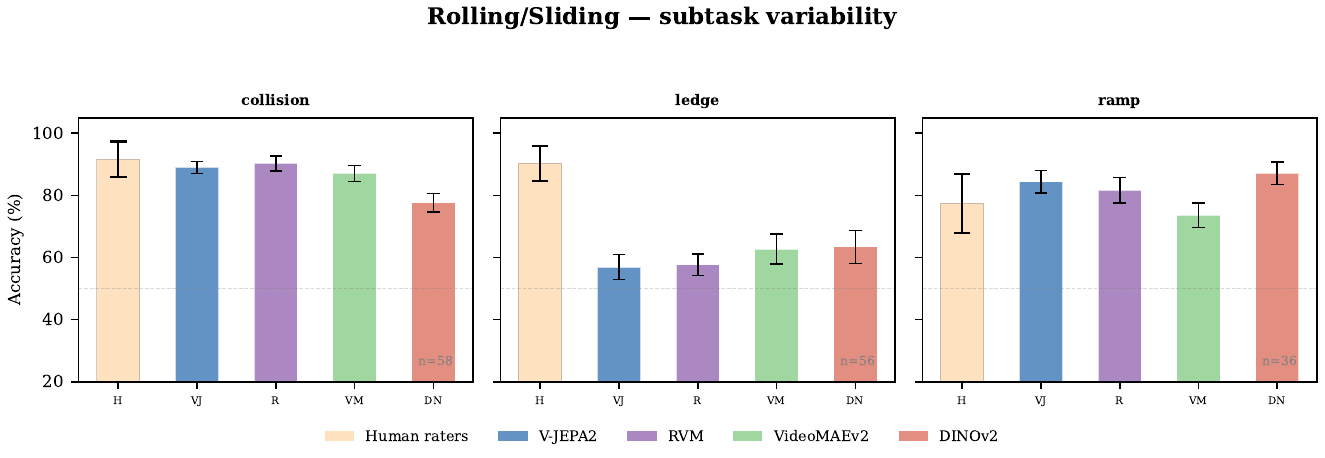}
\caption{Rolling/Sliding subtask variability.}
\label{fig:subtask_rolling}
\end{figure}

\begin{figure}[!htbp]
\centering
\includegraphics[width=\linewidth]{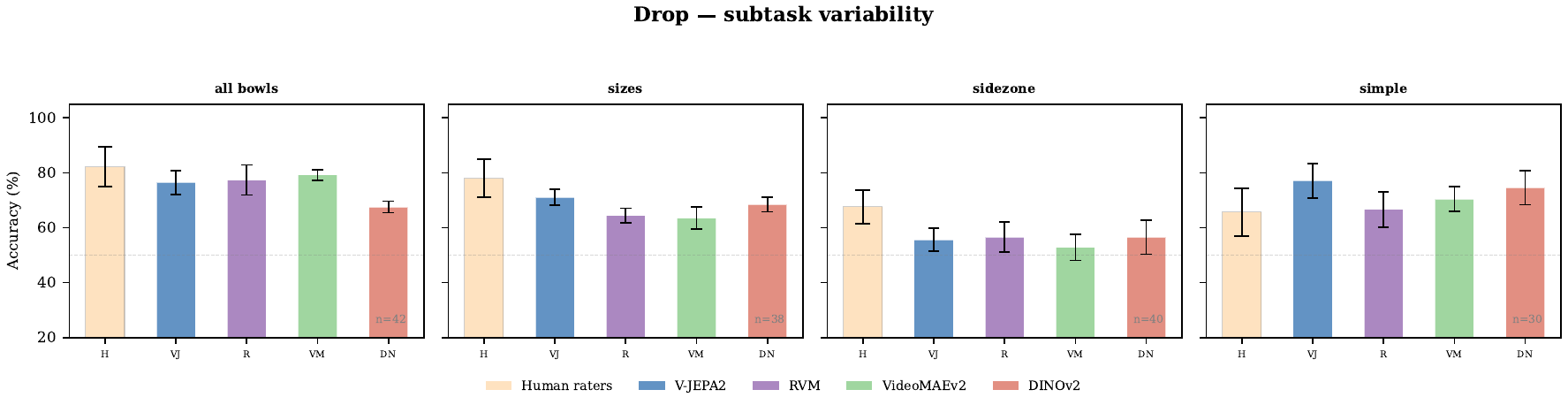}
\caption{Drop subtask variability.}
\label{fig:subtask_drop}
\end{figure}

\begin{figure}[!htbp]
\centering
\includegraphics[width=\linewidth]{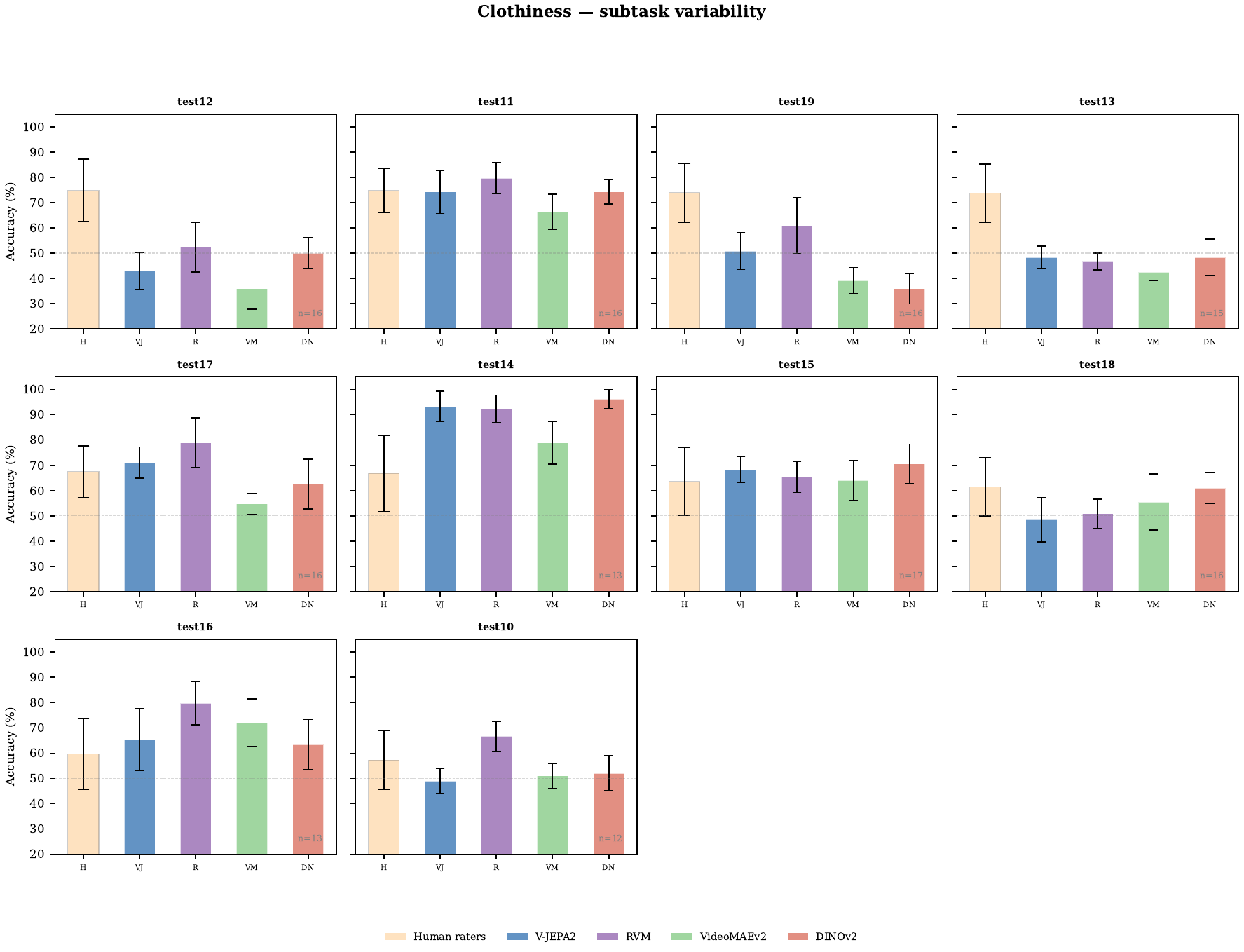}
\caption{Clothiness subtask variability.}
\label{fig:subtask_clothiness}
\end{figure}

\begin{figure}[!htbp]
\centering
\includegraphics[width=\linewidth]{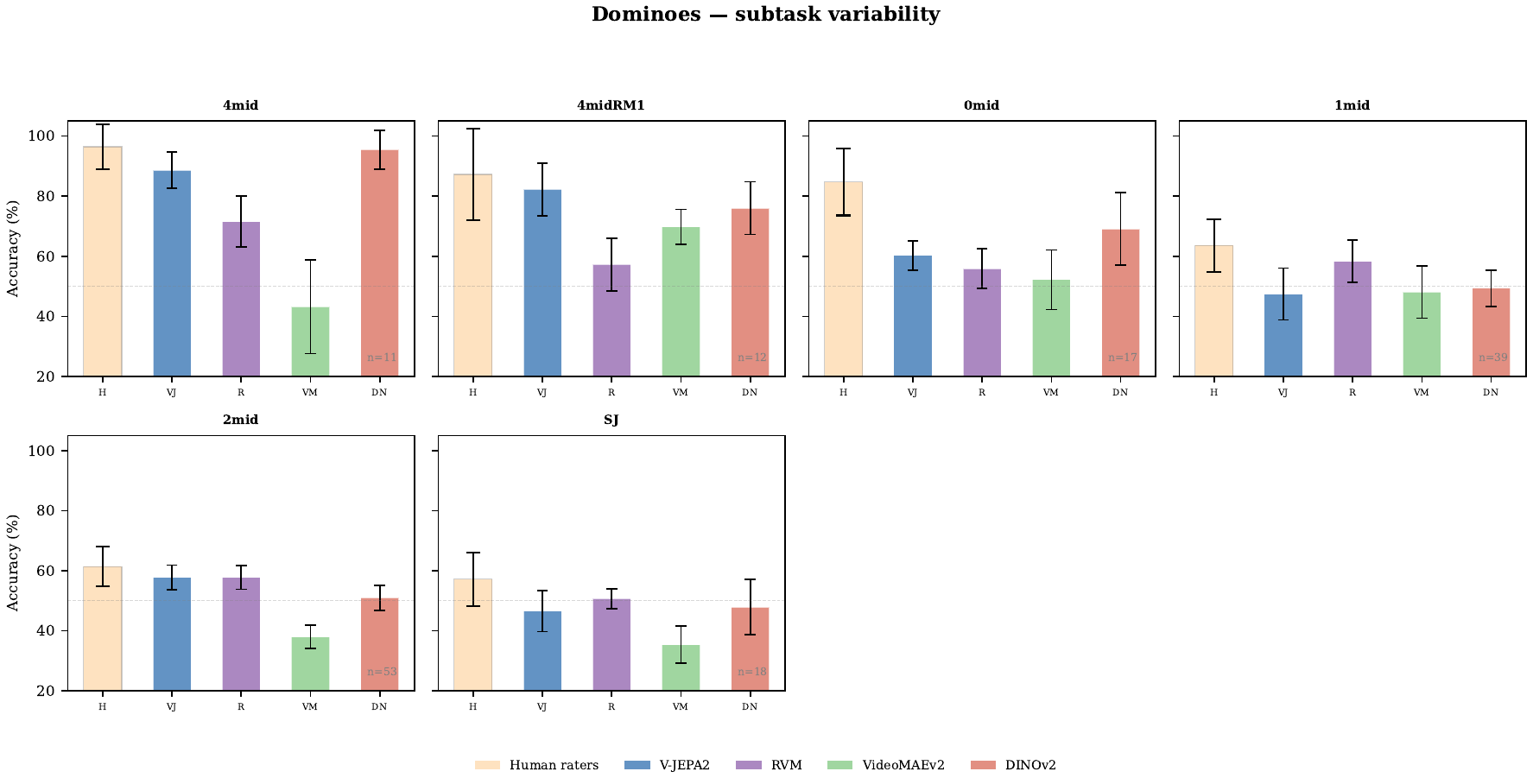}
\caption{Dominoes subtask variability.}
\label{fig:subtask_dominoes}
\end{figure}

\section{Calibration and Confidence Analysis}
\label{sec:calibration_appendix}

\begin{table}[!htbp]
\centering
\caption{Calibration across models. \vjepa{} is best-calibrated with fewest overconfident errors (model $>$80\% confident but wrong).}
\label{tab:calibration}
\small
\begin{tabular}{lccc}
\toprule
Model & ECE & High-conf.\ acc. & Overconf.\ errors \\
\midrule
\vjepa{}  & $\mathbf{0.166}$ & 84.2\% & 3 \\
\vmae{}   & 0.216 & 79.4\% & 16 \\
\dino{}   & 0.179 & 75.5\% & 24 \\
\bottomrule
\end{tabular}
\end{table}

\noindent\rvm{} is omitted from calibration analysis because its recurrent probe produces frame-level logits that are aggregated differently from the single-pass attentive probes used by the other three models, making ECE values not directly comparable.

\paragraph{Method.}
We compute Expected Calibration Error (ECE) by binning predicted probabilities into 10 equal-width bins and measuring the weighted average of $|\mathrm{acc}(b) - \mathrm{conf}(b)|$ across bins.
``High-confidence accuracy'' is accuracy restricted to samples where the model's predicted probability exceeds 0.8.
``Overconfident errors'' counts samples where predicted probability $>$0.8 but the prediction is wrong.

\paragraph{Per-scenario ECE.}
\vjepa{}'s calibration advantage is concentrated in contact-heavy scenarios: collision (ECE$\,{=}\,0.09$) and containment ($0.12$), where it achieves both high accuracy and well-calibrated confidence.
Clothiness shows the worst calibration across all models (ECE$\,{>}\,0.25$), consistent with its high uncertainty.

\paragraph{Cross-model error independence.}
Zero samples simultaneously fool all three models in this calibration analysis with high confidence, reinforcing that models fail on \emph{different} samples.
Mutual information analysis indicates that humans contribute $2{\times}$ the unique information about ground truth compared to any single model, supporting the three-level hierarchy where $\kappa_{\mathrm{HH}} \gg \kappa_{\mathrm{MM}} \gg \kappa_{\mathrm{MH}}$.

\section{Surprise Analysis}
\label{sec:surprise_appendix}

\noindent We focus on \vjepa{} as the highest-accuracy model; the same categorization applies to all models but \vjepa{} provides the clearest signal.

\paragraph{Category definitions.}
We define ``actively approaching'' as cutoff-frame approaching rate $>0$ (i.e., the target--zone distance is decreasing at the moment of truncation).
Combined with the ground-truth outcome (contact/no-contact), this yields four categories:
\emph{expected negative} (not approaching $\times$ no-contact, $n\!=\!356$),
\emph{surprise negative / near-miss} (approaching $\times$ no-contact, $n\!=\!244$),
\emph{expected positive} (approaching $\times$ contact, $n\!=\!305$), and
\emph{surprise positive / indirect causality} (not approaching $\times$ contact, $n\!=\!295$).

\paragraph{Results.}
\vjepa{} accuracy by category: expected negative $73.9\%$, surprise negative $79.9\%$, expected positive $74.8\%$, surprise positive $70.2\%$. The surprise-negative category is particularly revealing: despite correct ground truth being no-contact, objects are actively approaching at cutoff, yet \vjepa{} achieves its \emph{highest} category accuracy ($79.9\%$) while assigning low P(contact)$\,{=}\,0.24$, consistent with recognition of near-miss geometry. Conversely, surprise-positive samples (objects not approaching, but contact eventually occurs through indirect causation) yield the lowest accuracy ($70.2\%$), consistent with difficulty on indirect-causality cases.

\section{Overdispersion Analysis}
\label{sec:overdispersion_appendix}

\noindent We report \vjepa{} overdispersion as the primary model; all four models exhibit qualitatively similar overdispersion patterns. This diagnostic uses the submitted 8-seed probe set and is retained as a robustness analysis. The main paper retains the reviewed 8-seed cross-model protocol; separate 10-seed runs are reported only in the supplementary \rvm{} robustness section.

\paragraph{VIF calculation.}
For each sample $i$, we compute the observed variance of correctness across $K\!=\!8$ seeds: $\hat{v}_i = \frac{1}{K}\sum_k (c_{ik} - \bar{c}_i)^2$ where $c_{ik} \in \{0,1\}$ is seed $k$'s correctness. The expected variance under a Bernoulli model (independent seeds) is $v^{\mathrm{Bern}}_i = \bar{c}_i(1-\bar{c}_i)/K$. The variance inflation factor is VIF $= \langle \hat{v}_i \rangle / \langle v^{\mathrm{Bern}}_i \rangle$, averaged over all 1200 samples.

\paragraph{Results.}
A Bernoulli model predicts VIF$\,{=}\,1.0$. We observe VIF$\,{=}\,25.7$, indicating massive overdispersion: seeds share correlated errors driven by the frozen encoder's representation geometry.
Samples near the decision boundary in feature space are systematically unstable across \emph{all} probe initializations, confirming that inter-seed disagreement is a property of the \emph{representation}, not the probe.
The effect varies across scenarios: inter-seed agreement is strongest for collision ($\kappa\!=\!0.68$) and containment ($0.68$), and weakest for clothiness ($0.23$) and dominoes ($0.32$)---the two scenarios with the lowest absolute accuracy.
This suggests that the frozen encoder places difficult-scenario samples closer to the decision boundary, where small perturbations in probe initialization flip predictions.
Cross-model comparison reveals the same pattern: \vmae{} (VIF$\,{=}\,21.6$) and \dino{} (VIF$\,{=}\,16.1$) show lower overdispersion than \vjepa{} (VIF$\,{=}\,25.7$), likely because their lower overall accuracy concentrates more samples firmly on the wrong side of the boundary.

\section{Head Selection and Training Analysis}
\label{sec:head_analysis}

This head-selection audit uses the submitted 8-seed probe sweep. Of the 80 heads trained across 8 seeds (10 heads $\times$ 8 seeds), ${\sim}55\%$ collapse due to high learning rates (lr$\,{\geq}\,$3e-3).
This is expected behavior for attentive probes with depth-4 transformers: the search grid intentionally includes aggressive learning rates to ensure the optimum is interior to the grid rather than at its boundary.
All seeds retain 3--5 healthy heads out of 10, and the selected learning rate (1e-4) is consistent across seeds, confirming that the grid adequately covers the viable range.
The majority select weight decay$\,{=}\,$0.1. The oracle gap between train-accuracy selection and test-accuracy selection is only $0.36\%$, confirming that the standard protocol is near-optimal. Test accuracy plateaus at ${\sim}$epoch 10--12.

\section{Statistical Details}
\label{sec:statistical_details}

For every percentage we report Wilson score 95\% confidence intervals. Divergence gaps are tested via paired McNemar statistic with Holm--Bonferroni correction. The three-level agreement hierarchy (\S3.2 of the main text) uses 1,000 bootstrap iterations for $\kappa_{\mathrm{HH}}$ and $\kappa_{\mathrm{MM}}$; $95\%$ CIs are reported. Strategy fingerprint significance is assessed via permutation testing (10,000 permutations of sample labels).

\section{Entropy Decomposition and Four-Quadrant Analysis}
\label{sec:entropy_appendix}

We decompose per-sample uncertainty using binary entropy $H(p) = -p\log_2 p - (1{-}p)\log_2(1{-}p)$, computed separately for model seeds ($H_m$) and human raters ($H_h$). A threshold $\tau\!=\!0.811$ bits (corresponding to 25\%/75\% consensus) partitions samples into four quadrants:
\textbf{Q1}~(Easy): both populations confident ($H_m\!<\!\tau$, $H_h\!<\!\tau$);
\textbf{Q2}~(Model Blind): model uncertain, humans confident;
\textbf{Q3}~(Human Hard): model confident, humans uncertain;
\textbf{Q4}~(Genuinely Hard): both uncertain.

\begin{table}[!htbp]
\centering
\caption{Entropy quadrant distribution across models ($n\!=\!1200$). The asymmetry ratio $A\!=\!|$Q2$|/|$Q3$|$ quantifies whether model uncertainty misaligns with human uncertainty ($A\!=\!1$ would indicate symmetric uncertainty).}
\label{tab:entropy_quadrants}
\small
\begin{tabular}{lcccccc}
\toprule
Model & Q1 (Easy) & Q2 (M-Blind) & Q3 (H-Hard) & Q4 (Hard) & $A$ \\
\midrule
\vjepa{}  & 704 (58.7\%) & 103 (8.6\%) & 324 (27.0\%) & 69 (5.8\%) & 0.318 \\
\vmae{}   & 708 (59.0\%) & 99 (8.3\%) & 320 (26.7\%) & 73 (6.1\%) & 0.309 \\
\dino{}   & 682 (56.8\%) & 125 (10.4\%) & 318 (26.5\%) & 75 (6.3\%) & 0.393 \\
\bottomrule
\end{tabular}
\end{table}

The asymmetry ratio is consistently below one across the analyzed models ($A\!=\!0.31$--$0.39$; $A\!=\!0.318$ for \vjepa{}), indicating a structural imbalance: Q3 (model confident, humans uncertain) is about $3{\times}$ larger than Q2 (model uncertain, humans confident). This means models are confident on many samples where humans are uncertain---the pattern expected if models rely on behavioral cues that can be correct on average while diverging from human physical judgments.

Figure~\ref{fig:entropy_vjepa2} shows the entropy scatter plot for \vjepa{}: Q3 is dominated by clothiness (62 samples) and linking (63 samples), the two scenarios with the highest model--human disagreement, while Q1 is dominated by collision (155 samples), consistent with its low disagreement rate.

\begin{figure}[!htbp]
\centering
\includegraphics[width=0.7\linewidth]{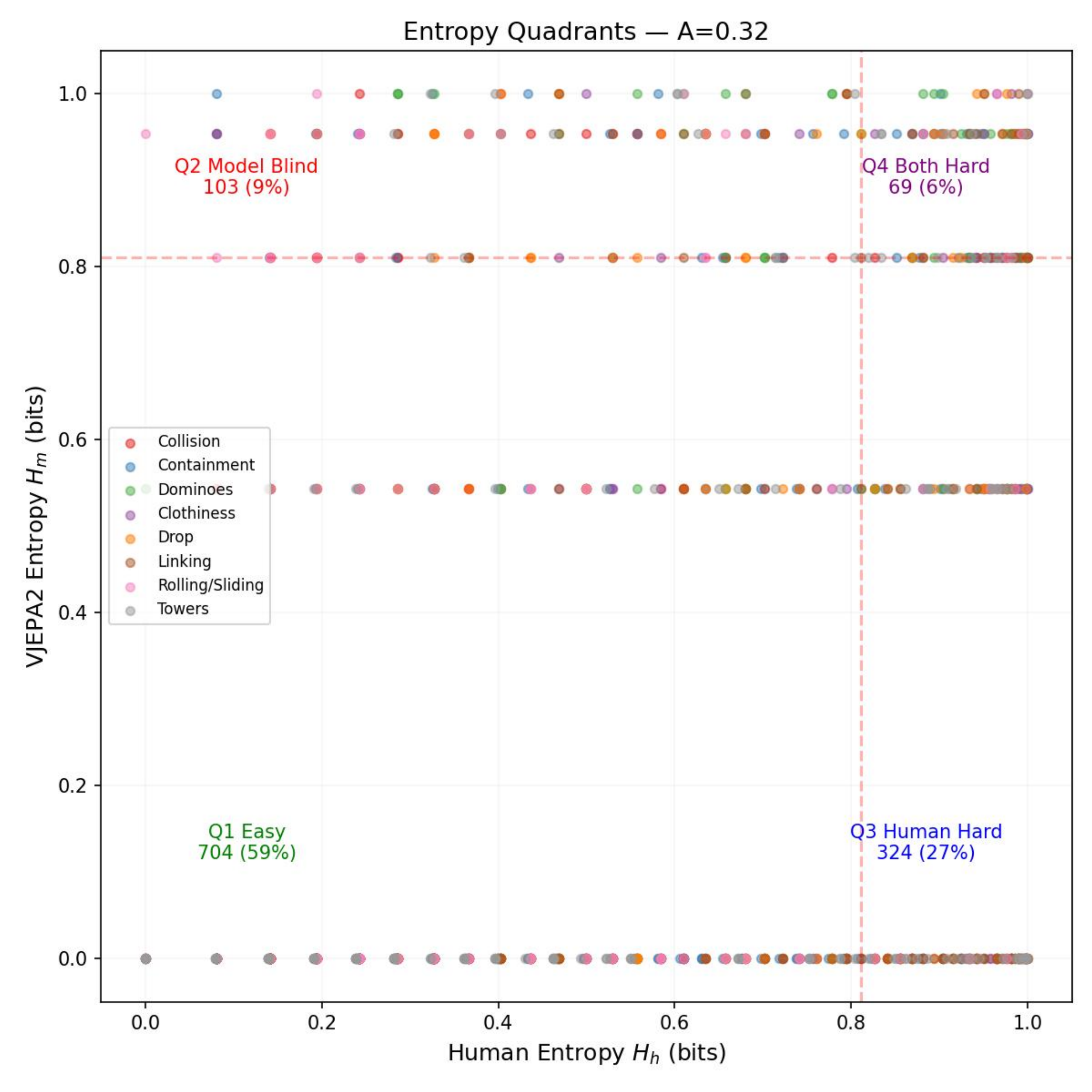}
\caption{\textbf{Entropy scatter plot (\vjepa{}).} Each point is a test sample; axes show model-seed entropy ($H_m$) and human-rater entropy ($H_h$). Dashed lines at $\tau\!=\!0.811$ bits partition the space into four quadrants. Q3 (bottom-right, model confident but humans uncertain) is $3{\times}$ larger than Q2 (top-left), reflecting systematic overconfidence on human-hard samples.}
\label{fig:entropy_vjepa2}
\end{figure}

\section{Physics Feature Attribution Details}
\label{sec:attribution_appendix}

This section provides full results for the 89-feature attribution analysis summarized in \S6 of the main text. We train Ridge regression, random forest (RF), and gradient boosting (GB) models via 5-fold cross-validation, predicting various divergence targets from physics features.

\noindent\textbf{Note on main-text numbers.}
The main text reports the best-performing regressor for each feature set to provide the most favorable test of each hypothesis: RF $R^2\!=\!{-}0.32$ for observable features (vs.\ Ridge $-0.77$, GB $-0.46$) and GB $R^2\!=\!{+}0.29$ for unobservable features (vs.\ Ridge $+0.10$, RF $+0.23$). This is a conservative choice: even the \emph{best} observable model has negative $R^2$, strengthening the null result under this feature set and protocol, while the unobservable result holds across all three regressors. Table~\ref{tab:attribution} below reports all methods for full transparency.

\begin{table}[!htbp]
\centering
\caption{Attribution results: observable features (A+B, 66 dim) vs.\ unobservable features (C, 23 dim). $R^2$ is cross-validated (negative $=$ worse than constant predictor). AUROC is for binary targets.}
\label{tab:attribution}
\small\setlength{\tabcolsep}{3pt}
\begin{tabular}{llccc}
\toprule
Target & Features & Ridge & RF & GB \\
\midrule
\multicolumn{5}{l}{\emph{$R^2$ (regression targets)}} \\
$|p_m - p_h|$ (divergence) & A+B (obs.) & $-0.77$ & $-0.32$ & $-0.46$ \\
$|p_m - p_h|$ (divergence) & C (unobs.) & $+0.10$ & $+0.23$ & $\mathbf{+0.29}$ \\
\midrule
\multicolumn{5}{l}{\emph{AUROC (classification targets)}} \\
Disagree binary             & A+B (obs.) & $0.57$ & $0.63$ & $0.61$ \\
Disagree binary             & C (unobs.) & $\mathbf{0.70}$ & $0.70$ & $0.70$ \\
Both-certain disagree       & A+B (obs.) & $0.56$ & $0.69$ & $0.63$ \\
Both-certain disagree       & C (unobs.) & $0.80$ & $\mathbf{0.85}$ & $0.85$ \\
Human correct               & A+B (obs.) & $0.68$ & $0.68$ & $0.65$ \\
Human correct               & C (unobs.) & $0.80$ & $0.81$ & $\mathbf{0.83}$ \\
\bottomrule
\end{tabular}
\end{table}

\paragraph{Key findings.}
Observable features (A+B) consistently fail to predict model--human divergence ($R^2\!<\!0$), while unobservable outcome features~(C) achieve positive $R^2$ and substantially higher AUROC across all targets.
The strongest signal emerges for ``both-certain disagree'' samples---cases where model and human are both confident but give opposite answers.
Here, unobservable features reach AUROC$\,{=}\,0.85$, indicating that the complexity of the \emph{unobserved} future (collision chains, settling time, contact flips) strongly predicts where models and humans diverge.
This supports the main-text interpretation that divergence is more strongly associated with unobserved future outcomes than with directly visible scene features in our feature-attribution analysis.

\section{Feature Set Correspondence}
\label{sec:feature_correspondence}

Table~\ref{tab:feature_correspondence} lists the 37-feature core subset used for strategy fingerprinting (\S3.4 of the main text) and its correspondence with the 89-feature systematic set used for attribution (\S6 of the main text). Of 37 features, 32 have direct semantic equivalents in the 89-feature set.
The 5 unmatched features (marked ``---'') are camera-relative, temporal-derivative, or frame-index quantities that were excluded from the systematic 89-feature extraction to avoid redundancy with their physical counterparts.

\begin{table}[H]
\centering
\caption{Correspondence between 37-feature (strategy fingerprint) and 89-feature (attribution) sets. Match type: E\,=\,exact, P\,=\,partial (similar but not identical computation), ---\,=\,no equivalent.}
\label{tab:feature_correspondence}
\scriptsize\setlength{\tabcolsep}{2.5pt}
\begin{tabular}{llll}
\toprule
37-dim name & Ly & 89-dim equivalent & Mt \\
\midrule
\multicolumn{4}{l}{\emph{Layer A: Static/Scene (6 features)}} \\
\texttt{n\_distractors}          & A & \texttt{A03\_n\_distractors}         & E \\
\texttt{n\_occluders}            & A & \texttt{A04\_n\_occluders}           & E \\
\texttt{n\_physical\_objects}    & A & \texttt{A02\_n\_physical\_objects}   & E \\
\texttt{target\_mass}            & A & \texttt{A05\_target\_mass}           & E \\
\texttt{target\_friction}        & A & \texttt{A07\_target\_dynamic\_friction} & P \\
\texttt{target\_bounciness}      & A & \texttt{A06\_target\_bounciness}     & E \\
\midrule
\multicolumn{4}{l}{\emph{Layer B: Observable Dynamics (24 features)}} \\
\texttt{target\_total\_displacement} & B & \texttt{B07\_obs\_target\_displacement} & E \\
\texttt{target\_max\_speed}      & B & \texttt{B02\_obs\_target\_speed\_max}   & E \\
\texttt{target\_mean\_speed}     & B & \texttt{B01\_obs\_target\_speed\_mean}  & E \\
\texttt{target\_final\_speed}    & B & \texttt{B03\_obs\_target\_speed\_final} & E \\
\texttt{target\_motion\_onset\_pct} & B & ---                                  & --- \\
\texttt{total\_kinetic\_energy\_mean} & B & \texttt{B33\_obs\_scene\_ke\_mean}  & E \\
\texttt{total\_kinetic\_energy\_max}  & B & \texttt{B34\_obs\_scene\_ke\_max}   & E \\
\texttt{obs\_collision\_count}   & B & \texttt{B23\_obs\_collision\_count}    & E \\
\texttt{obs\_target\_collision\_count} & B & \texttt{B27\_obs\_target\_collision\_count} & E \\
\texttt{target\_distance\_from\_camera} & B & ---                              & --- \\
\texttt{speed\_trajectory\_slope}  & B & ---                                  & --- \\
\texttt{speed\_trajectory\_var}  & B & \texttt{B04\_obs\_target\_speed\_variance} & E \\
\texttt{ke\_trajectory\_slope}   & B & \texttt{B36\_obs\_scene\_ke\_trend}   & P \\
\texttt{ke\_drop\_ratio}         & B & ---                                    & --- \\
\texttt{cutoff\_target\_speed}   & B & \texttt{B03\_obs\_target\_speed\_final} & E \\
\texttt{cutoff\_target\_ke}      & B & \texttt{B35\_obs\_scene\_ke\_final}   & P \\
\texttt{cutoff\_pair\_distance}  & B & \texttt{B17\_obs\_tz\_distance\_final} & E \\
\texttt{cutoff\_approaching\_rate} & B & \texttt{B20\_obs\_tz\_approaching\_rate} & E \\
\texttt{cutoff\_scene\_ke}       & B & \texttt{B35\_obs\_scene\_ke\_final}   & E \\
\texttt{cutoff\_target\_moving}  & B & \texttt{B13\_obs\_target\_has\_moved} & E \\
\texttt{contact\_at\_cutoff}     & B & \texttt{B40\_obs\_contact\_at\_cutoff} & E \\
\texttt{min\_pair\_distance}     & B & \texttt{B18\_obs\_tz\_distance\_min}  & E \\
\texttt{min\_dist\_frame}        & B & ---                                    & --- \\
\texttt{n\_direction\_changes}   & B & \texttt{B10\_obs\_target\_dir\_changes} & E \\
\midrule
\multicolumn{4}{l}{\emph{Layer C: Unobservable Outcome (7 features)}} \\
\texttt{prediction\_horizon}     & C & \texttt{C12\_prediction\_horizon}     & E \\
\texttt{outcome\_frame}          & C & \texttt{C10\_outcome\_determ\_frame}  & E \\
\texttt{late\_contact}           & C & \texttt{C13\_late\_contact}           & E \\
\texttt{contact\_flips}          & C & \texttt{C19\_future\_contact\_flips}  & E \\
\texttt{settling\_time}          & C & \texttt{C14\_settling\_time}          & E \\
\texttt{min\_dist\_after\_cutoff} & C & \texttt{C17\_future\_tz\_distance\_min} & E \\
\texttt{post\_cutoff\_surprise}  & C & \texttt{C23\_outcome\_surprise}       & E \\
\bottomrule
\end{tabular}
\end{table}

\section{Subtask Heatmap}
\label{sec:subtask_heatmap}

\begin{figure}[!htbp]
\centering
\includegraphics[width=0.82\linewidth]{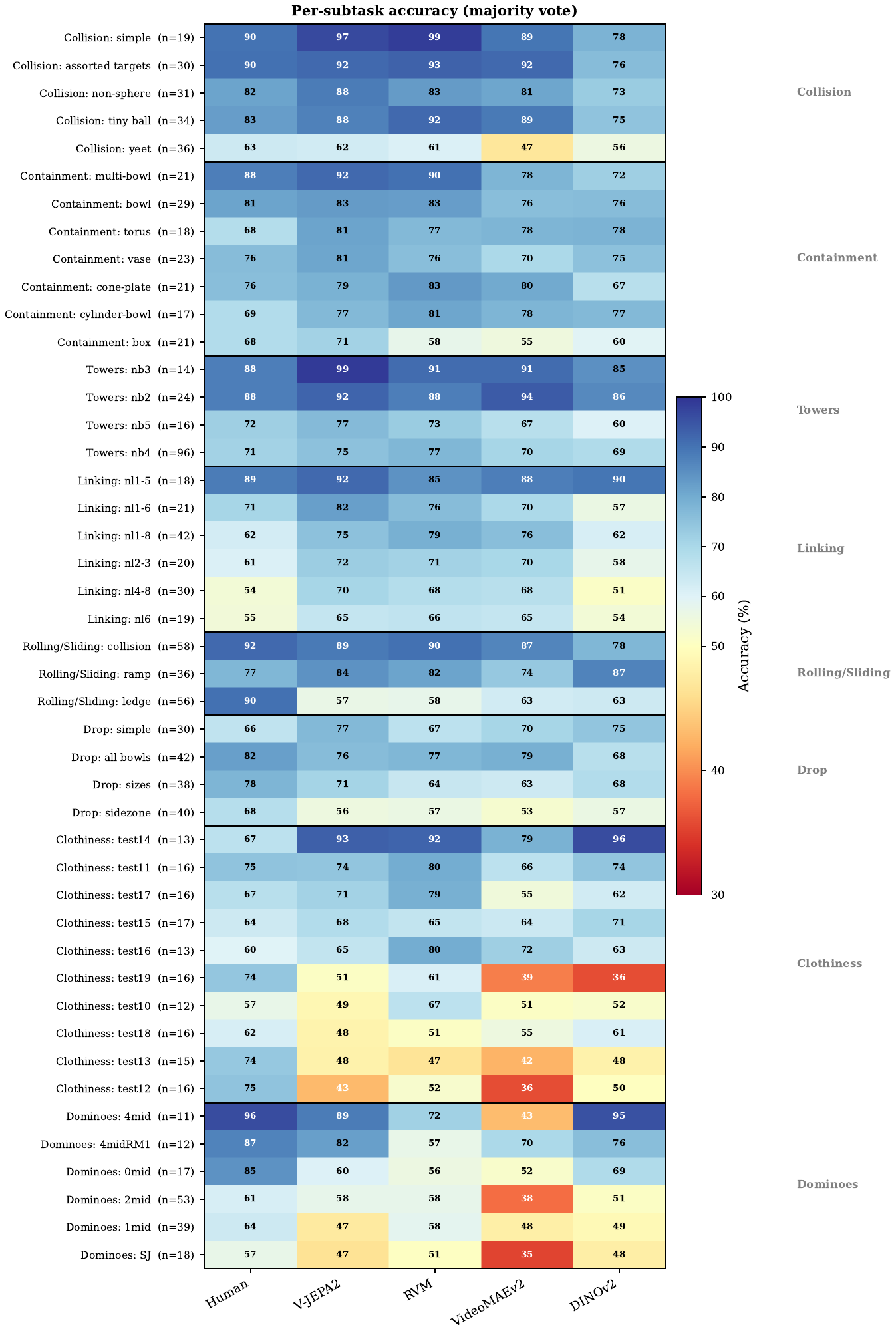}
\caption{Per-subtask accuracy heatmap (45 subtasks $\times$ 5 agents). Rows grouped by scenario, sorted by \vjepa{} accuracy. Color: red ($<$50\%) $\to$ yellow (50\%) $\to$ blue ($>$80\%).}
\label{fig:subtask_heatmap}
\end{figure}

\noindent The heatmap reveals fine-grained cross-model differences.
Largest gap: Dominoes 4-Mid shows a 52pp difference between \dino{} ($95\%$) and \vmae{} ($43\%$).
Notable rank reversals include Dominoes 4-Mid (\dino{} $>$ \vmae{}) and Drop all-bowls (\vmae{} $79\%$ $>$ \vjepa{} $76\%$).

\clearpage

\section{Qualitative High-Confidence Disagreement Cases}
\label{sec:qualitative_cases}

Figure~\ref{fig:case_study} shows representative both-certain disagreement cases selected from real Physion trials in response to reviewer requests for qualitative inspection. Each row shows the start frame, the 1.5s observation cutoff, the full outcome, and the contact-response rates for human raters and the evaluated model populations. The examples cover both directions of model--human reversal and are intended as qualitative anchors for the aggregate disagreement statistics, rather than as a mechanistic decomposition of why each stimulus is difficult.

\begin{figure}[!htbp]
\centering
\includegraphics[width=0.81\linewidth]{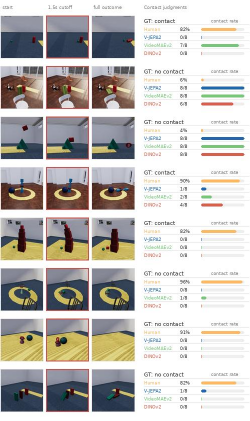}
\caption{\textbf{Qualitative high-confidence disagreement cases.} Rows show representative stimuli where human raters and model seeds give high-confidence opposite contact predictions. From left to right, each row shows the start frame, the observation cutoff, the full outcome, and contact-response rates for humans, \vjepa{}, \vmae{}, and \dino{}. The first five rows are human-majority-correct/model-wrong cases; the last three rows are model-correct/human-majority-wrong cases.}
\label{fig:case_study}
\end{figure}

\fi

\end{document}